\documentclass[11pt, a4paper, twocolumn, copyright, gdm]{google}

\usepackage[authoryear, sort&compress, round]{natbib}
\uselogo{} 

\usepackage{algorithm}
\usepackage{algorithmic}

\usepackage{booktabs}

\usepackage{subcaption}
\usepackage{chessboard}
\usepackage{physics}

\usepackage{graphicx}
\usepackage{xurl}
\usepackage{xspace}

\title{Conditional Generation of Creative Chess Puzzles with Diffusion Models}

\author[1]{Aatu Selkee}
\author[1]{Severi Rissanen}
\author[2]{Xidong Feng}
\author[2]{Tom Zahavy}
\author[1]{Eric Malmi}

\affil[1]{Aalto University}
\affil[2]{\thepa{}{}}

\newcommand{\vone}{\textit{v1}\xspace}
\newcommand{\vtwo}{\textit{v2}\xspace}

\cbDefinePgfFieldStyle{masktoken}{
    \fill[gray!80] (-0.5,-0.5) rectangle (0.5,0.5);
    \node[text=white, font=\sffamily\bfseries\Large] at (0,0) {?};
}

\begin{abstract}
While modern language models demonstrate impressive generative capabilities, they often struggle with constrained, counter-intuitive creative tasks. To address this limitation, we explore chess puzzle generation as a rigorous testbed for computational creativity and reasoning, a domain where altering a single piece can invalidate an entire solution. We propose a novel approach for conditional generation of creative chess puzzles using masked diffusion models. Unlike previous methods, our non-directional diffusion approach allows for conditioning on specific tactical themes and partial board positions. We introduce a novel auxiliary task of simultaneous best-move prediction, which improves solution uniqueness by 11.6\% and theme-conditioning accuracy by 2.5\%. To further optimize solution uniqueness and theme conditioning, we establish a reinforcement learning framework adapted from Denoising Diffusion Policy Optimization (DDPO). This RL training increases the yield of unique and theme-matching positions by 89.1\%. Finally, we release the first open-weights models (Appendix~\ref{sec:model weights and training}) for chess puzzle generation, offering a new pathway for controllable, creative generation.
\end{abstract}

\begin{document}

\maketitle

\section{Introduction}

The idea of machine intelligence has intrigued researchers for decades, dating back to Turing's early work on indistinguishable machine behavior \citep{turing1950computingmachineryandintelligence}. In recent years, this pursuit has culminated in the rapid rise of Large Language Models (LLMs), which \citet{jones2026largelanguagemodelspassastandardthreepartyturingtest} argue are already passing the standard Turing test. A key benchmark for evaluating the success of these systems is creativity \citep{colton2012computationalcreativity}. However, while modern language models are capable of creative work, such as creating art, writing stories or making lyrics, they struggle to come up with truly new and unique items \citep{wenger2026llmsarehomogenouslycreative}. To address this limitation, we attempt to enhance the controllability of creative generation with a small, domain-specific language model.

\begin{figure}[t]
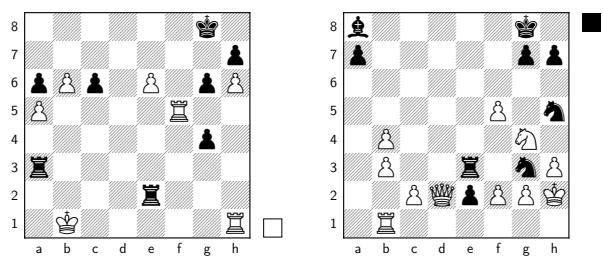

    \begin{subfigure}[t]{0.23\textwidth}
        \centering
        \resizebox{\textwidth}{!}{
            \chessboard[
                setfen=6k1/7p/pPp1P1pP/P4R2/6p1/r7/4r3/1K5R w - - 0 43
            ]
        }
        \caption{\textit{Themes:} advantage, short, endgame, rook endgame.} % \\ \textit{Solution:} Rb5 cxb5 b7}
        \label{fig:final puzzle1}
    \end{subfigure}
    \hfill
    \begin{subfigure}[t]{0.23\textwidth}
        \centering
        \resizebox{\textwidth}{!}{
            \chessboard[
                setfen=b5k1/p5pp/8/5P1n/1P4N1/1P2r1nP/2PQpPPK/1R6 b - - 1 38
            ]
        }
        \caption{\textit{Themes:} fork, under promotion, promotion, endgame} %\\
        % \textit{Solution:} Nf1+ Rxf1 exf1=N+ Kg1 Nxd2}
        \label{fig:final puzzle2}
    \end{subfigure}
    \caption{Example chess puzzles generated by the proposed model. The themes are provided as input to the model. Solutions: (a) Rb5 cxb5 b7 (b) Nf1+ Rxf1 exf1=N+ Kg1 Nxd2}
    \label{fig:final puzzles}
\end{figure}

Although the most commonly used language models today are autoregressive, discrete diffusion models have recently been successfully adapted to discrete sequences and language modeling tasks \citep{austin2021structureddenoisingdiffusionmodels,nie2025largelanguagediffusionmodels,ye2025dream7bdiffusionlarge}. While autoregressive generation builds sequences token-by-token, diffusion models iteratively refine the entire sequence globally \citep{shi2025simplifiedgeneralizedmaskeddiffusion}. Autoregressive models have generally been outperforming diffusion models with respect to output quality, but recent works demonstrate competitive results using diffusion models \citep{nie2025largelanguagediffusionmodels,ye2025dream7bdiffusionlarge}. Moreover, they can yield significant speedups, offering compelling quality--inference speed tradeoffs enabled by parallel decoding \citep{wu2026fastdllm}. Given these benefits, we select a masked diffusion model for our creative generation task.

Evaluating the creative abilities of language models is an open question, and previous work has turned to chess puzzle generation as a contained playground \citep{feng2025generatingcreativechesspuzzles}. We employ this same framework as it allows us to concentrate on the controllability of creative generation. This domain requires the models to balance between creativity and reasoning, as chess puzzles are generally delicate. Changing even one piece on the board may entirely change the puzzle to something completely different.

To train our model for this task, we first employ supervised training on the Lichess puzzles dataset \citep{lichesspuzzledatabase}. Our model takes a list of themes as input and generates positions, which match the given themes. After supervised training, we observe that around $22.37\%$ of the positions generated by our model include a unique solution and the correct themes. After supervised training, we employ reinforcement learning, allowing us to nearly double the rate of positions with unique solution and correct themes to $42.32\%$. Moreover, given the small model size (268M parameters), we can efficiently generate a large number of positions and filter out the ones not passing the criteria.

The main contributions of this work are the following:
\begin{enumerate}
    \item We demonstrate the effectiveness of masked diffusion models for creative chess puzzle generation. In contrast to previous work, that has primarily focused on autoregressive models \citep{feng2025generatingcreativechesspuzzles}, the proposed diffusion model provides the user a significantly greater control over the creative process by supporting conditioning of the puzzle generation on ($i$) a set of themes, ($ii$) a partially specified position, and ($iii$) the best move. This unlocks new practical possibilities such as practicing specific types of puzzles that a user wants to improve on.
    \item We propose a novel auxiliary task of predicting the best move together with generating the puzzle. We show that this improves the solution uniqueness and accuracy of theme-conditioning by $11.6\%$ and $2.5\%$, respectively. Additionally it allows us to prevent policy collapse during RL.
    \item We introduce an effective RL recipe on top of masked diffusion models, which increases the rate of unique and thematically correct positions by $89.1\%$. Getting RL to work effectively is challenging with language models in general and it is less explored on top of discrete diffusion models.
    \item We release the first open weights models for chess puzzle generation at: \url{https://github.com/naapeli/Chess-Puzzle-Generation}
\end{enumerate}

Example puzzles generated by our model are presented in Figure \ref{fig:final puzzles}. For each puzzle, the solution and the themes on which we conditioned the generation are in the caption. These puzzles were hand-picked from the generated puzzles based on the elegance of the solution.

\section{Related work}

\paragraph{Machine intelligence and creativity}

Evaluating and expanding machine intelligence through computational creativity has long served as a key benchmark for artificial intelligence systems~\citep{turing1950computingmachineryandintelligence, colton2012computationalcreativity}. While modern Generative AI models demonstrate strong capabilities across unstructured tasks like text and image generation, they often exhibit limited capacity for open-ended discovery or generating truly novel, highly constrained, counter-intuitive artifacts~\citep{wenger2026llmsarehomogenouslycreative}. To evaluate and improve generative models beyond standard tasks, researchers increasingly investigate specialized domains requiring both aesthetic quality and tight computational constraints.

To address these limitations, recent work explores pairing generative architectures with evolutionary search and expert evaluation in specialized domains. For instance, \citet{earle2026search} substituted human selectors with vision-language models (VLMs) in the canonical Picbreeder framework to test open-ended semantic discovery, while \citet{banarse2026evolution} integrated genetic algorithms with foundation models to co-evolve complex 3D forms. Beyond purely visual media, AI pipelines have been applied to constrained spatial tasks like flat-foldable origami design~\citep{zahavy2026corigami}, and recent empirical evaluations confirm that expert reviewers frequently rate machine-generated creative outputs as highly novel and counter-intuitive compared to human baselines~\citep{veeriah2025evaluating}.

\paragraph{Chess puzzle generation}

Although chess puzzle generation is a niche research field, it has been studied quite extensively despite its limited practical applicability. Traditional ways of generating chess puzzles include evolutionary search approaches, as well as iterative approaches. These algorithms iteratively modify one or a set of chess positions with a goal of making better puzzles \citep{feng2025generatingcreativechesspuzzles,iqbal2021acomputationalmethodofoptimizingchesscompositions,iqbal2017computercomposesfabledproblem}. Both algorithms use a fitness function that assesses qualities similar to our reward function \eqref{eq:reward function}. One more modern approach is to use deep learning for this task \citep{feng2025generatingcreativechesspuzzles}. \citeauthor{feng2025generatingcreativechesspuzzles} train a model that generates chess puzzles. They manage to train a model that successfully beats a baseline of the Lichess puzzles dataset \citep{lichesspuzzledatabase} in terms of counter-intuitivity. We do not position our work as attempting to beat their baseline in puzzle quality metrics. Instead, we attempt to add additional features useful for actual use, such as conditional generation. Additionally, an open source implementation allows anyone to use our model in practice.

\paragraph{Masked diffusion models}

While autoregressive large language models (LLMs) currently dominate language modelling, discrete diffusion models have emerged as a compelling alternative \citep{nie2025largelanguagediffusionmodels,ye2025dream7bdiffusionlarge}. Unlike autoregressive generation, which generates sequences token-by-token in a causal manner, discrete diffusion models iteratively refine the entire sequence globally using bidirectional attention \citep{austin2021structureddenoisingdiffusionmodels}. This non-causal structure allows the model to leverage context from the entire input sequence simultaneously and has been shown to reduce training data memorization \citep{luo2026characterizingmemorizationdiffusionlanguage}. Furthermore, diffusion models offer variable inference-time efficiency by adjusting the number of denoising steps \citep{austin2021structureddenoisingdiffusionmodels}. However, they are computationally more expensive to pretrain to achieve comparable text quality \citep{nie2025scalingmaskeddiffusionmodels}. Some tasks, which do not have the inherent left-to-right structure of language, such as sudoku might even benefit from the approach of diffusion models, where the order of the generated tokens is not fixed \citep{Ye2025beyondautoregression}.

\paragraph{Reinforcement learning with diffusion models}

In recent years, great effort has been put into improving reinforcement learning for diffusion models \citep{ou2025principledrldiffusionllms,zhao2025d1scalingreasoningdiffusion,wang2026d2improvingresoning,rojas2025improvingreasoningdiffusionlanguage,black2024trainingdiffusionmodelsreinforcement,fan2023dpok}. Compared to traditional LLMs, diffusion models suffer from computationally intractable sequence-level likelihoods. More traditional methods solve this by evaluating the likelihoods step-wise, which requires many forward passes \citep{black2024trainingdiffusionmodelsreinforcement, fan2023dpok}, whereas other algorithms approximate the likelihoods with one forward pass \citep{ou2025principledrldiffusionllms,zhao2025d1scalingreasoningdiffusion,wang2026d2improvingresoning,rojas2025improvingreasoningdiffusionlanguage}. In this research, we initially attempted to use the ELBO-based sequence-level policy optimization (ESPO) algorithm \citep{ou2025principledrldiffusionllms}, however, we were unable to find a working configuration. Afterwards, we turned to denoising diffusion policy optimization (DDPO), which we found to perform better despite being less efficient in theory.

\section{Methods}
\label{sec:methods}

\subsection{Masked diffusion models}

We frame our approach within the masked diffusion paradigm, where sequences are generated by iteratively choosing tokens from a vocabulary to replace mask tokens $[\mathtt{MASK}]$ \citep{shi2025simplifiedgeneralizedmaskeddiffusion}. The forward process corrupts an unmasked sequence $y$ into a noisy state $y_t$ at time $t \in [0, 1]$ via $q(y_t \mid y) = \text{Cat}(y_t; \bar{Q}(t)^\top y)$. Using a masking schedule $\alpha_t$, the transition matrix $\bar{Q}(t) = \alpha_t I + (1 - \alpha_t)\mathbf{1}e_m^\top$ transitions sequences from a fully unmasked to a completely masked as time increases \citep{shi2025simplifiedgeneralizedmaskeddiffusion}.

To reverse this corruption, a denoising network $\mu_\theta(y_t, x)$ is introduced. We employ a transformer encoder architecture with 268M parameters based on the 270M transformer from \citet{ruoss2024amortizedplanninglargescaletransformers} as the denoising network. We also employ cross attention to incorporate theme conditioning and SwiGLU (with a widening factor of 2.66 instead of 4) \citep{shazeer2020gluvariantsimprovetransformer}. The transition to a less masked state $y_s$ ($s < t$) is defined as:
\begin{equation}
    p_\theta(y_s \mid y_t, x) = \text{Cat}(y_s; \bar{R}^{\mu_\theta(y_t, x)}(t, s)^\top y_t),
    \label{eq:reverse process}
\end{equation}
where $\bar{R}$ is the reverse transition matrix \citep{shi2025simplifiedgeneralizedmaskeddiffusion}. Conceptually, a token masked at step $t$ is unmasked at step $s$ with probability $\frac{\alpha_s - \alpha_t}{1 - \alpha_t}$, using a token sampled from the network's output.
To represent the board in a machine-readable format, we pad the Forsyth-Edwards Notation (FEN) to a fixed length and convert it into tokens. Other implementation details are provided in Table~\ref{tab:hyperparameters} in the Appendix.

\subsection{Supervised training}

During supervised training, we optimize the evidence lower bound (ELBO) on the Lichess puzzle dataset \citep{shi2025simplifiedgeneralizedmaskeddiffusion}:
\begin{equation}
    \begin{split}
        &\mathcal{L}_\theta(y \mid x) =\\
        &-\int_0^1 \frac{\alpha_t'}{1-\alpha_t} \mathbb{E}_{q(y_t \mid y)} \Bigg[\sum_{i=1}^L \mathbb{1}[y_t^{(i)} = [\mathtt{MASK}]]\\
        &(y^{(i)})^\top \log \mu_\theta(y_t, x)^{(i)} \Bigg] \mathrm{d}t,
    \end{split}
    \label{eq:elbo}
\end{equation}
where $L$ is the sequence length, $x$ is the conditioning information, and $y_t \sim q(y_t \mid y)$ is the corrupted sequence. To evaluate the integral and the expectation, we employ a 1-point Monte-Carlo (MC) estimate. A comprehensive list of all hyperparameters used during training can be found in Table \ref{tab:hyperparameters} in the Appendix. Following previous methodology, we select the checkpoint with the lowest validation loss, which occurs at 1,000,000 steps \citep{feng2025generatingcreativechesspuzzles}.

To evaluate how best move prediction affects model performance, we train two different configurations: one that generates both the puzzle and the best move, and another that only generates the puzzle. In addition to these configurations, we also experiment with block diffusion to investigate the effect of best move prediction on puzzle quality. With block diffusion \citep{arriola2025blockdiffusion}, we split the generative process into two phases: The first phase generates the board state while artificially keeping the best move tokens masked. Second, after all the board tokens have been generated, we sample the best move. We use this as a middle ground between the two standard models, as the block diffusion model has learned internal representations for the best move but is restricted from explicitly unmasking those tokens during board generation.

\subsection{Denoising Diffusion Policy Optimization}
Similar to \citet{feng2025generatingcreativechesspuzzles}, we apply reinforcement learning to optimize the pretrained model for puzzle quality metrics. For RL training, we adapt the Denoising Diffusion Policy Optimization (DDPO) algorithm \citep{black2024trainingdiffusionmodelsreinforcement}. Although DDPO was originally developed for image-generation models, we adapt it to masked diffusion models by using the transition probabilities from the reverse process \eqref{eq:reverse process} as the policy $\pi_\theta(a_t \mid s_t, x) = p_\theta(a_t \mid s_t, x)$. Additionally, we adapt the group relative advantage calculation from GRPO \citep{shao2024deepseekmathpushinglimitsmathematical}, which is originally not part of DDPO. DDPO acts in a step-wise Markov Decision Process (MDP), where the states are defined to be partially masked token sequences $s_t$. The actions $a_t$ correspond to sampling a new partially masked token sequence with the transitions being deterministic. The reward is given at the end of the trajectory for the full generated sequence and propagated through the trajectory via discounting.

As shown by \citet{feng2025generatingcreativechesspuzzles}, a pure reward maximization procedure suffers from entropy collapse. Like their work, we employ diversity filtering mechanisms along with an entropy bonus and a KL divergence penalty to prevent entropy collapse and ensure the realism of the puzzles. However, we optimize these penalties directly via gradient descent by incorporating them into the loss function. Using this setup, we seek to maximize the following objective:
\begin{equation}
    \begin{split}
        \mathcal{J}&(\theta) = \\
        &L(\theta) - \beta \mathbb{E}_t\left[\mathbb{KL}(\pi_\theta \parallel \pi_{\text{ref}})\right] + \gamma \mathbb{E}_t\left[H(\pi_\theta)\right]
    \end{split}
\end{equation}
where $L(\theta) = \mathbb{E}_t [ \min(\rho_t \hat{A}_t, \text{clip}(\rho_t, 1-\epsilon, 1+\epsilon)\hat{A}_t) ]$ is the proximal policy optimization (PPO) objective \citep{schulman2017proximalpolicyoptimizationalgorithms}, and $\mathbb{E}_t\left[\mathbb{KL}(\pi_\theta \parallel \pi_{\text{ref}})\right]$ and $\mathbb{E}_t\left[H(\pi_\theta)\right]$ are the expected step-wise KL divergence and the entropy, respectively. The advantages $\hat{A}_t$ are computed and normalized group-wise using an unbiased Monte Carlo estimate proposed by \citet{kool2019unbiasedbaseline}.

\subsection{Detecting chess puzzles}
Following \citet{feng2025generatingcreativechesspuzzles}, we define a chess position to be a puzzle if it passes legality, uniqueness \eqref{eq:uniqueness_condition}, and counter-intuitiveness \eqref{eq:counter-intuitiveness_condition} criteria. We attempt to define and implement the metrics in the same way as \citet{feng2025generatingcreativechesspuzzles}, but some differences remain in the distance calculations.

\paragraph{Uniqueness}
A position $s$ is unique if the winning chances after the best move are considerably larger than those after the second best move \citep{lichesspuzzler,feng2025generatingcreativechesspuzzles}.
\begin{equation}
    \mathbb{I}_{\text{uni}}(s) = w(a_{\text{best}} \mid s) - w(a_{\text{second}} \mid s) > \tau_{\text{uni}},
    \label{eq:uniqueness_condition}
\end{equation}
where $w(a \mid s)\footnote{$w(a \mid s) = 2 / (1 + \exp(-0.00368208\cdot \mathrm{centipawns})) - 1$}\in[-1, 1]$ can be interpreted as the chance of winning after move $a$ in position $s$ \citep{Lichessaccuracymetric} and $\tau_{\text{uni}} = 0.5$.

\paragraph{Counter-intuitiveness}
The counter-intuitiveness value of a position is defined as follows \citep{feng2025generatingcreativechesspuzzles}:
\begin{equation}
    r_{\text{cnt}}(s) = 0.8 \cdot \frac{d_{\text{cp}}}{50} - 0.1 \cdot \frac{v_{\text{captured}}}{9}
    \label{eq:counter-intuitiveness_condition}
\end{equation}
where $d_{\text{cp}} \le 50$ is the critical depth at which Stockfish first discovers the best move, and $v_{\text{captured}}$ is the value of the potentially captured piece. A position is considered counter-intuitive if $r_{\text{cnt}}(s) > \tau_{\text{cnt}} = 0.1$.

\paragraph{Reward function}
We define the RL reward function $R(s)$ as:
\begin{equation}
    R(s) = \begin{cases}
        -2                                 & \text{if } \neg\mathbb{I}_{\text{legal}}(s)                                                                  \\
        1 + 20\cdot r_{\text{cnt}}(s) & \text{if } \mathbb{I}_{\text{legal}}(s) \cdot \mathbb{I}_{\text{filter}}(s) \\
        0                                  & \text{otherwise}
    \end{cases}
    \label{eq:reward function}
\end{equation}
where $\mathbb{I}_{\text{filter}}(s) = \mathbb{I}_{\text{uni}}(s) \cdot \mathbb{I}_{\text{diversity}}(s) \cdot \mathbb{I}_{\text{pieces}}(s) \cdot \mathbb{I}_{\text{theme}}(s) \cdot \mathbb{I}_{\text{move}}(s)$. $\mathbb{I}_{\text{pieces}}(s)$ filters out positions in which the piece count of some piece exceeds that of the starting position, $\mathbb{I}_{\text{theme}}(s)$ indicates that the generated puzzle adheres the requested themes, and $\mathbb{I}_{\text{move}}(s)$ forces the predicted best move to match the actual best move in the generated position. The diversity indicator factorizes as $\mathbb{I}_{\text{diversity}}(s) = \mathbb{I}_{\text{intra}}(s) \cdot \mathbb{I}_{\text{inter}}(s) \cdot \mathbb{I}_{\text{opponent}}(s)$, where $\mathbb{I}_{\text{intra}}(s)$ filters out positions that are too similar to each other in the same batch, $\mathbb{I}_{\text{inter}}(s)$ filters out positions that are too similar to positions in previous batches, and $\mathbb{I}_{\text{opponent}}(s)$ ensures the opponent responses to the best moves are sufficiently different. Both intra- and inter-batch distances are computed using the Levenshtein distance \citep{li2007normalizedlevenshteindistance} over the FEN and the principal variation (PV). A position passes the conditions if the minimum distance to all other positions is greater than a threshold. For more details, we direct the reader to \citet{feng2025generatingcreativechesspuzzles}, as apart from the theme and move conditioning, the PV distance and the exact form of the reward function, our diversity filtering setup is very similar to theirs.

\section{Experimental evaluation}
\label{sec:experiments}

Our experimental evaluation is structured as follows. We first analyze how best move prediction affects the puzzle generation process. We also compare our model to the ones trained by \citet{feng2025generatingcreativechesspuzzles}. Second, we evaluate the reinforcement learning phase. Finally, we present how conditioning affects the generation, specifically during the RL phase.

\begin{table*}[t]
    \centering
    \caption{Puzzle quality metrics for different model configurations. All models were sampled with a temperature of 1.0 and 256 denoising steps. Errors represent 95\% confidence intervals. Counter-intuitiveness is computed over positions with unique solutions.}
    \label{tab:metrics}
    \resizebox{\textwidth}{!}{%
        \begin{tabular}{llcccccc}
            \toprule
            Training & Predict best move & Legal (\%) & Unique (\%) & Counter-Intuitive (\%) & Puzzle (\%) & Unique$^{*}$ \& Theme (\%) & Theme | Unique$^{*}$ (\%) \\
            \midrule
            Supervised & No & 98.97 $\pm$ 0.05 & 39.16 $\pm$ 0.23 & 1.00 $\pm$ 0.08 & 0.39 $\pm$ 0.03 & 19.10 $\pm$ 0.19 & 82.04 $\pm$ 0.39 \\
            Supervised & Block & 98.99 $\pm$ 0.05 & 41.25 $\pm$ 0.24 & 0.99 $\pm$ 0.07 & 0.41 $\pm$ 0.03 & 20.77 $\pm$ 0.20 & 83.64 $\pm$ 0.36 \\
            Supervised & Yes & 98.96 $\pm$ 0.05 & 43.69 $\pm$ 0.24 & 0.93 $\pm$ 0.07 & 0.41 $\pm$ 0.03 & 22.37 $\pm$ 0.20 & 84.05 $\pm$ 0.34 \\
            RL \vone & Yes & 99.76 $\pm$ 0.04 & 38.02 $\pm$ 0.43 & \textbf{2.23 $\pm$ 0.20} & \textbf{0.85 $\pm$ 0.08} & 16.40 $\pm$ 0.32 & 73.80 $\pm$ 0.82 \\
            RL \vtwo & Yes & \textbf{99.94 $\pm$ 0.01} & \textbf{66.15 $\pm$ 0.20} & 0.36 $\pm$ 0.03 & 0.23 $\pm$ 0.02 & \textbf{42.32 $\pm$ 0.22} & \textbf{87.21 $\pm$ 0.21} \\
            \midrule
            \multicolumn{2}{l}{Supervised \citep{feng2025generatingcreativechesspuzzles}} & 99.72 & 30.89 & 1.11 & 0.34 & — & — \\
            \midrule
            \multicolumn{2}{l}{\textit{Lichess puzzles dataset}} & 100.00 $\pm$ 0.00 & 95.05 $\pm$ 0.10 & 2.33 $\pm$ 0.07 & 2.22 $\pm$ 0.06 & 74.66 $\pm$ 0.19 & 94.57 $\pm$ 0.11\\
            \bottomrule
        \end{tabular}
    }
    \par\smallskip
    \begin{minipage}{\textwidth}
        \footnotesize
        $^{*}$ The uniqueness criterion used for the Unique \& Theme (\%) and Theme | Unique (\%) metrics includes small differences, such as the player cannot be losing in the generated position or the solution cannot be one move long unless it is a mate-in-one. It is more strict than the criterion used for Unique (\%), which uses \eqref{eq:uniqueness_condition}.
    \end{minipage}
\end{table*}

\begin{table*}[t]
    \centering
    \caption{Self-distance and Lichess-distance metrics after supervised training and RL. Metrics are computed over positions passing the uniqueness filter. We align the number of positions with \citet{feng2025generatingcreativechesspuzzles} and use 100k samples for Lichess-distances and 40k samples for the self-distances. Despite this, only board distances are comparable to \citet{feng2025generatingcreativechesspuzzles} due to differing implementations of the PV distance.}
    \label{tab:distances}
    \begin{tabular}{llcccc}
        \toprule
        Training & Predict best move & Board (Self) & Board (Lichess) & PV (Self) & PV (Lichess) \\
        \midrule
        Supervised & No & 11.63 $\pm$ 0.01 & 11.16 $\pm$ 0.01 & 0.78 $\pm$ 0.002 & 0.93  $\pm$ 0.001 \\
        Supervised & Block & 11.67 $\pm$ 0.01 & 11.18 $\pm$ 0.01 & 0.80 $\pm$ 0.002 & 0.93 $\pm$ 0.001 \\
        Supervised & Yes & 11.70 $\pm$ 0.01 & 11.19 $\pm$ 0.01 & 0.81 $\pm$ 0.002 & 0.93 $\pm$ 0.001 \\
        RL \vone & Yes & \textbf{15.15 $\pm$ 0.01} & \textbf{14.34 $\pm$ 0.01} & 0.76 $\pm$ 0.002 & \textbf{0.97 $\pm$ 0.001} \\
        RL \vtwo & Yes & 12.12 $\pm$ 0.01 & 12.10 $\pm$ 0.01 & \textbf{0.88 $\pm$ 0.002} & \textbf{0.97 $\pm$ 0.001} \\
        \midrule
        \multicolumn{2}{l}{Supervised \citep{feng2025generatingcreativechesspuzzles}} & 10.859 & 8.528 & (0.837) & (0.637) \\
        \midrule
        \multicolumn{2}{l}{\textit{Lichess puzzles dataset}} & 11.83 $\pm$ 0.01 & — & 0.95 $\pm$ 0.001 & — \\
        \bottomrule
    \end{tabular}
\end{table*}

\subsection{Best move prediction improves puzzle metrics}

Table~\ref{tab:metrics} summarizes the puzzle metrics of the models trained with supervised training. Interestingly, we find that simultaneously generating the best move alongside the board state leads to measurable improvements in both uniqueness and theme match rates compared to generating the board alone or predicting the best move with block diffusion. The model that predicts the best move has a 11.6\% higher uniqueness rate than the model that does not predict the best move. Additionally, the theme match rate assuming a unique solution is increased by 2.5\%. Although the highest puzzle rate is shared by the models that generate the best move and the highest counter-intuitiveness by the model with no best move prediction, the confidence intervals clearly overlap making the best model inconclusive in terms of these metrics. Across all supervised configurations, the theme match rate remains high indicating that the model successfully aligns its generations with the requested conditioning in the majority of cases and reflecting a solid grasp of varied tactical concepts. Despite this, there remains some themes that the model is unable to understand well. These themes include for instance castling and under promotion. All such themes are rare even in the Lichess puzzles dataset making the pretraining not sufficient for such themes.

To quantify the novelty and diversity of the generated positions, we utilize the distance metrics summarized in Table~\ref{tab:distances}. For the baseline Lichess puzzle dataset, the board distance is 11.83, which corresponds to a difference of roughly 12 distinct board elements between a puzzle and the most similar other position. The Lichess dataset exhibits a baseline PV distance of 0.95, representing a high diversity of solutions where the start or target squares of the moves typically differ by at least one rank or file. As shown in Table~\ref{tab:distances}, all three supervised configurations closely preserve these distances. The board self-distances are only slightly lower than that of the original Lichess dataset, while the PV self-distance remains also quite similar. Furthermore, the board and PV distances to the Lichess dataset itself are high. Importantly, the choice of the best-move generation strategy has virtually no impact on these diversity metrics, demonstrating that coupling chess understanding with the generation process to improve puzzle quality does not compromise diversity.

Comparing our model to the one trained by \citet{feng2025generatingcreativechesspuzzles} in Table~\ref{tab:metrics}, we observe that after supervised training, our models have higher uniqueness and puzzle rates. In contrast, their model has similar legal rates as ours after RL. Additionally, their counter-intuitiveness is slightly higher than ours. It is easy, however, to increase the puzzle metrics at the cost of diversity. In Table~\ref{tab:distances}, we observe that both board distances are considerably higher for our models compared to theirs. Notably, we have not managed to align the PV distance implementation to theirs. Therefore, the PV distances between our models and theirs are not comparable. However, overall, based on the board distance and the puzzle metrics, conditioning the generation on themes helps the model perform better in practice.

\subsection{Reinforcement learning}

\begin{figure*}[t]
    \centering
    \begin{subfigure}[t]{0.24\textwidth}
        \centering
        \includegraphics[width=\textwidth]{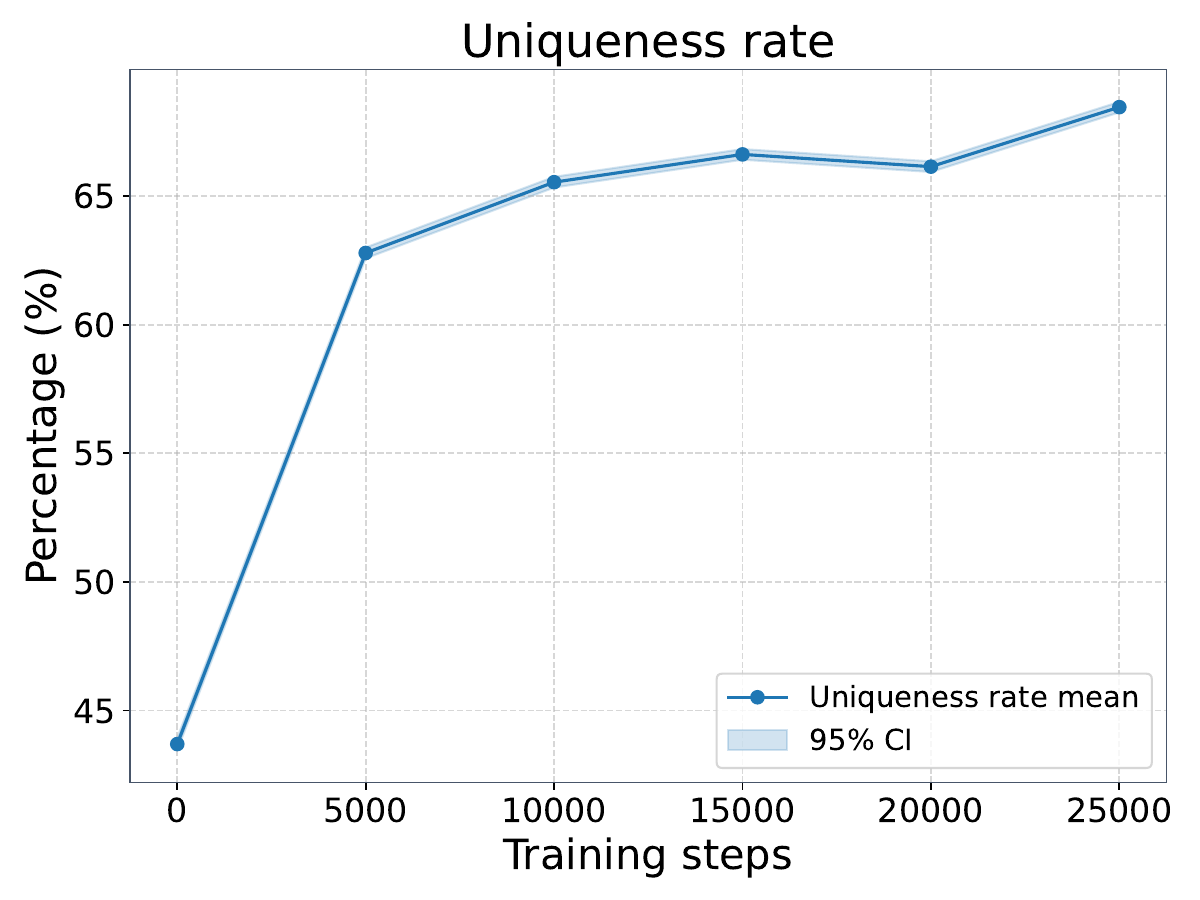}
        \caption{Uniqueness rate}
        \label{fig:rl_uniqueness}
    \end{subfigure}
    \hfill
    \begin{subfigure}[t]{0.24\textwidth}
        \centering
        \includegraphics[width=\textwidth]{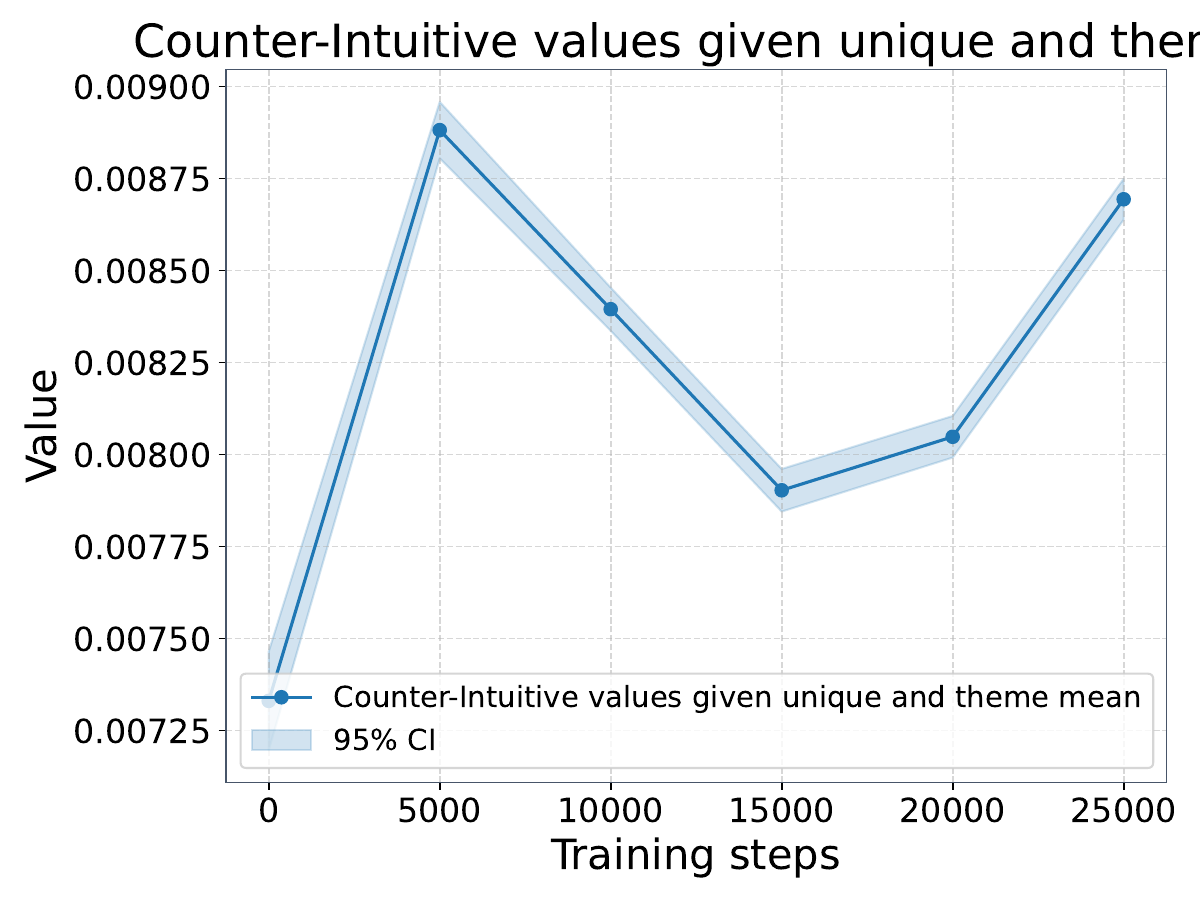}
        \caption{Counter-intuitive values}
        \label{fig:rl_counter_intuitive}
    \end{subfigure}
    \hfill
    \begin{subfigure}[t]{0.24\textwidth}
        \centering
        \includegraphics[width=\textwidth]{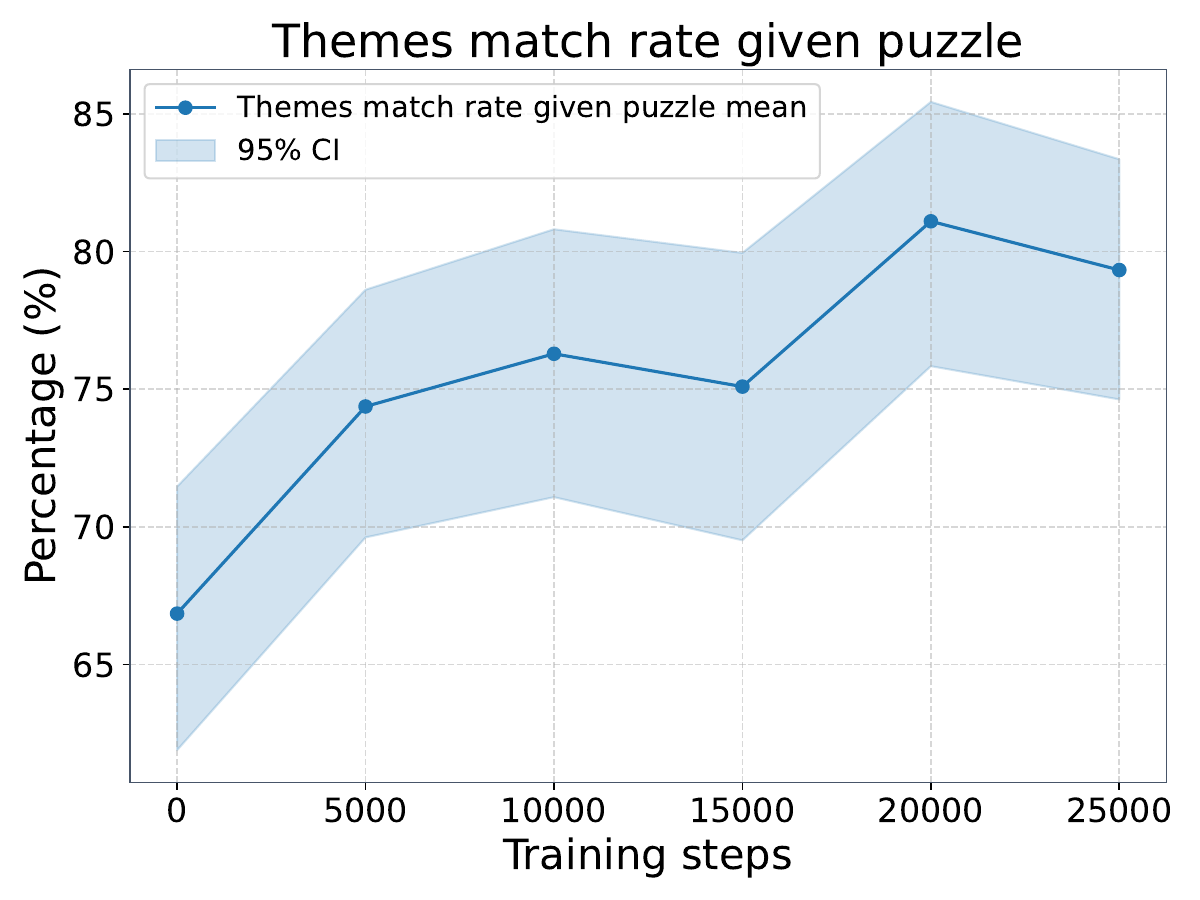}
        \caption{Themes match rate}
        \label{fig:rl_themes}
    \end{subfigure}
    \hfill
    \begin{subfigure}[t]{0.24\textwidth}
        \centering
        \includegraphics[width=\textwidth]{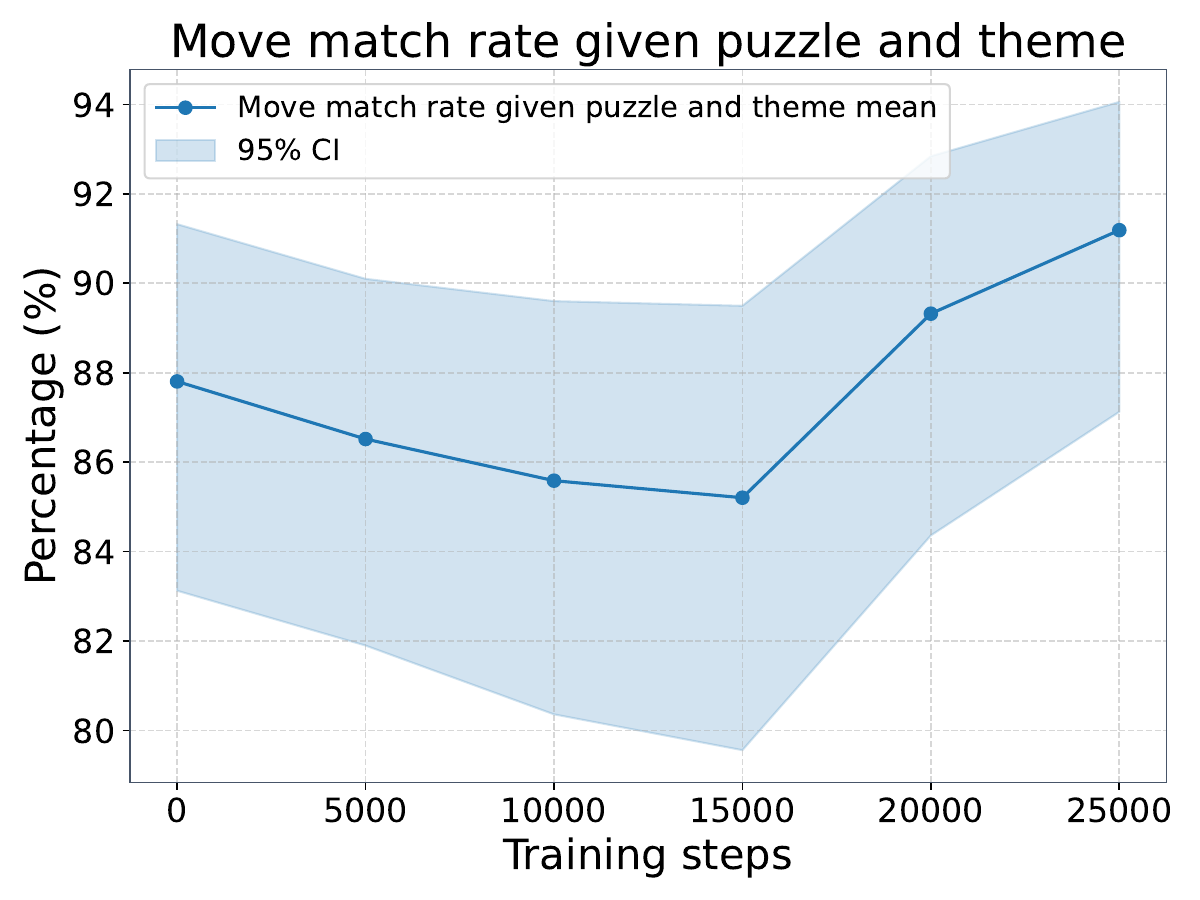}
        \caption{Move match rate}
        \label{fig:rl_move_match}
    \end{subfigure}

    \vspace{0.2cm}

    \begin{subfigure}[t]{0.24\textwidth}
        \centering
        \includegraphics[width=\textwidth]{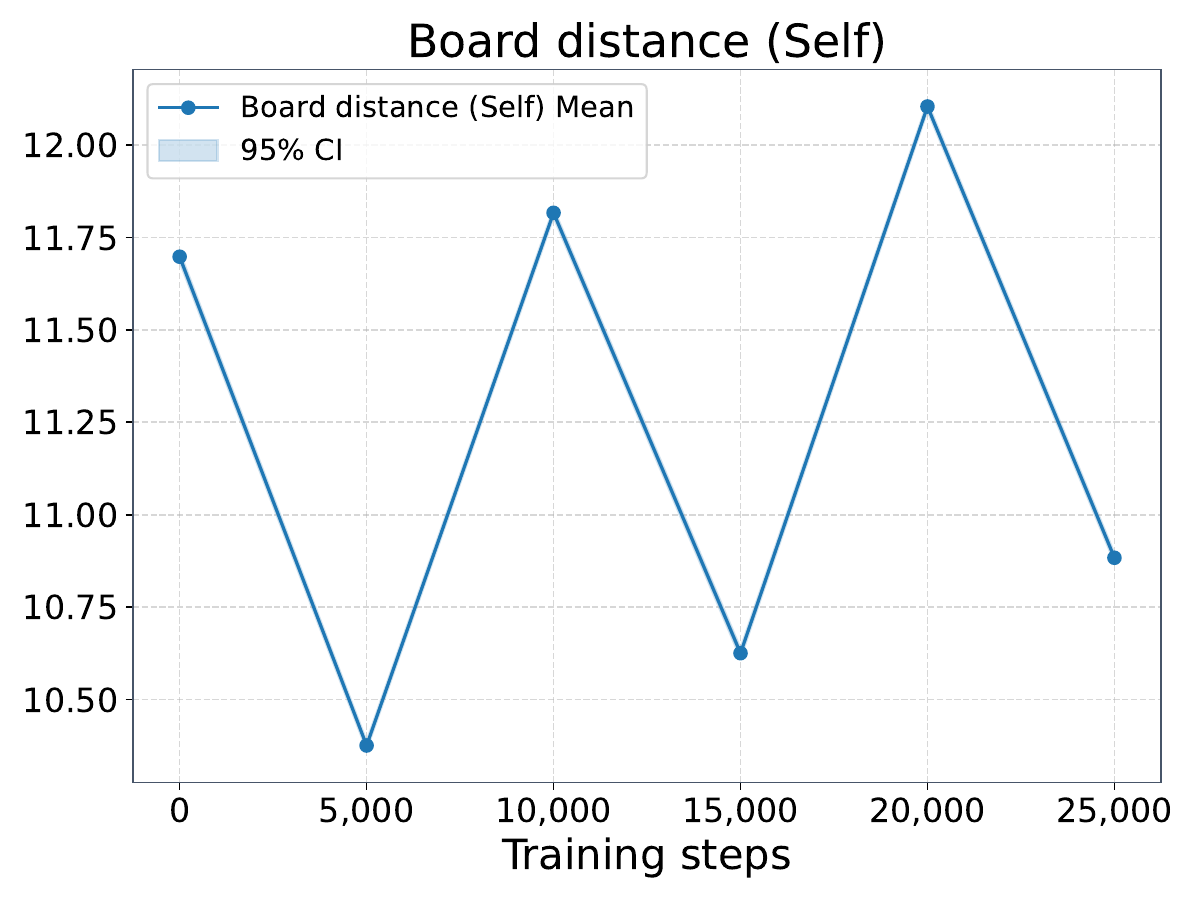}
        \caption{Board distance (Self)}
        \label{fig:rl_self_board}
    \end{subfigure}
    \hfill
    \begin{subfigure}[t]{0.24\textwidth}
        \centering
        \includegraphics[width=\textwidth]{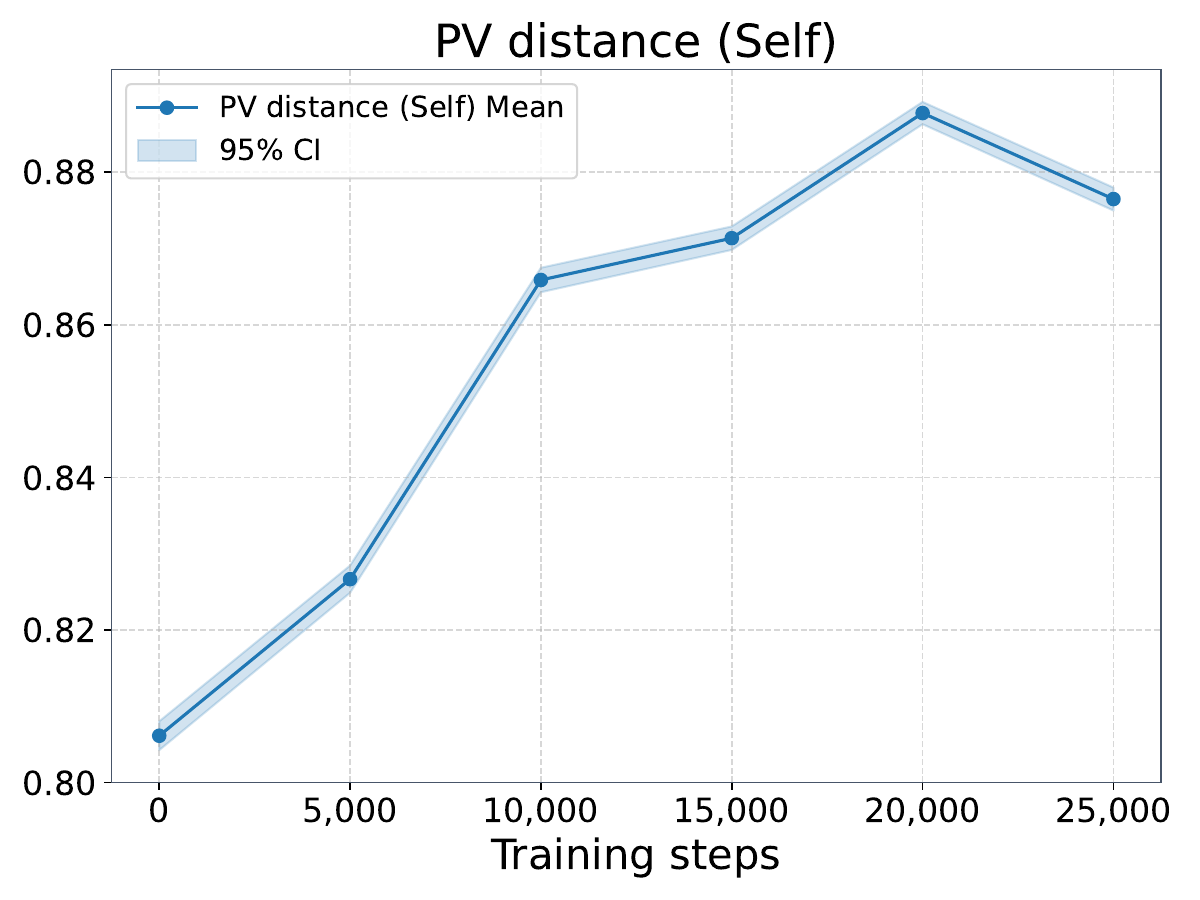}
        \caption{PV distance (Self)}
        \label{fig:rl_self_pv}
    \end{subfigure}
    \hfill
    \begin{subfigure}[t]{0.24\textwidth}
        \centering
        \includegraphics[width=\textwidth]{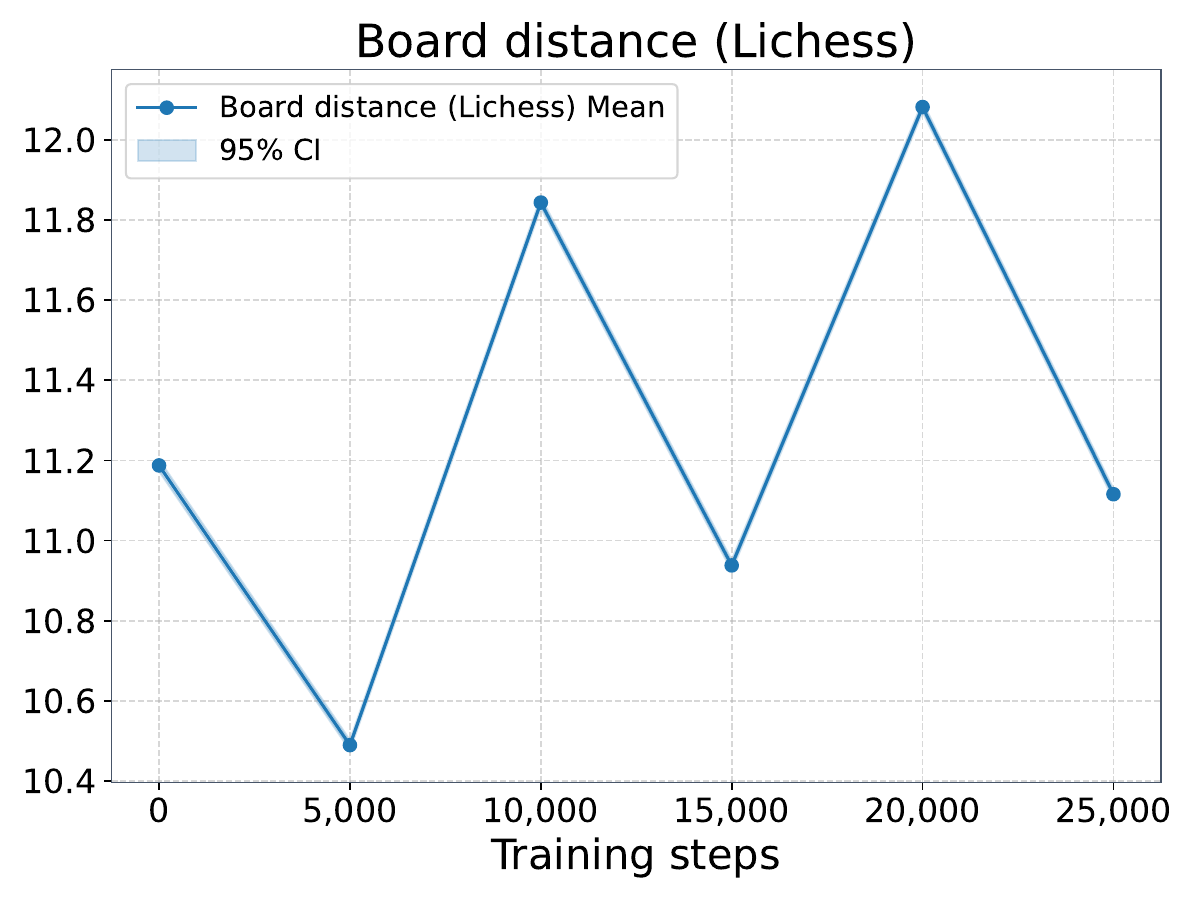}
        \caption{Board distance (Lichess)}
        \label{fig:rl_lichess_board}
    \end{subfigure}
    \hfill
    \begin{subfigure}[t]{0.24\textwidth}
        \centering
        \includegraphics[width=\textwidth]{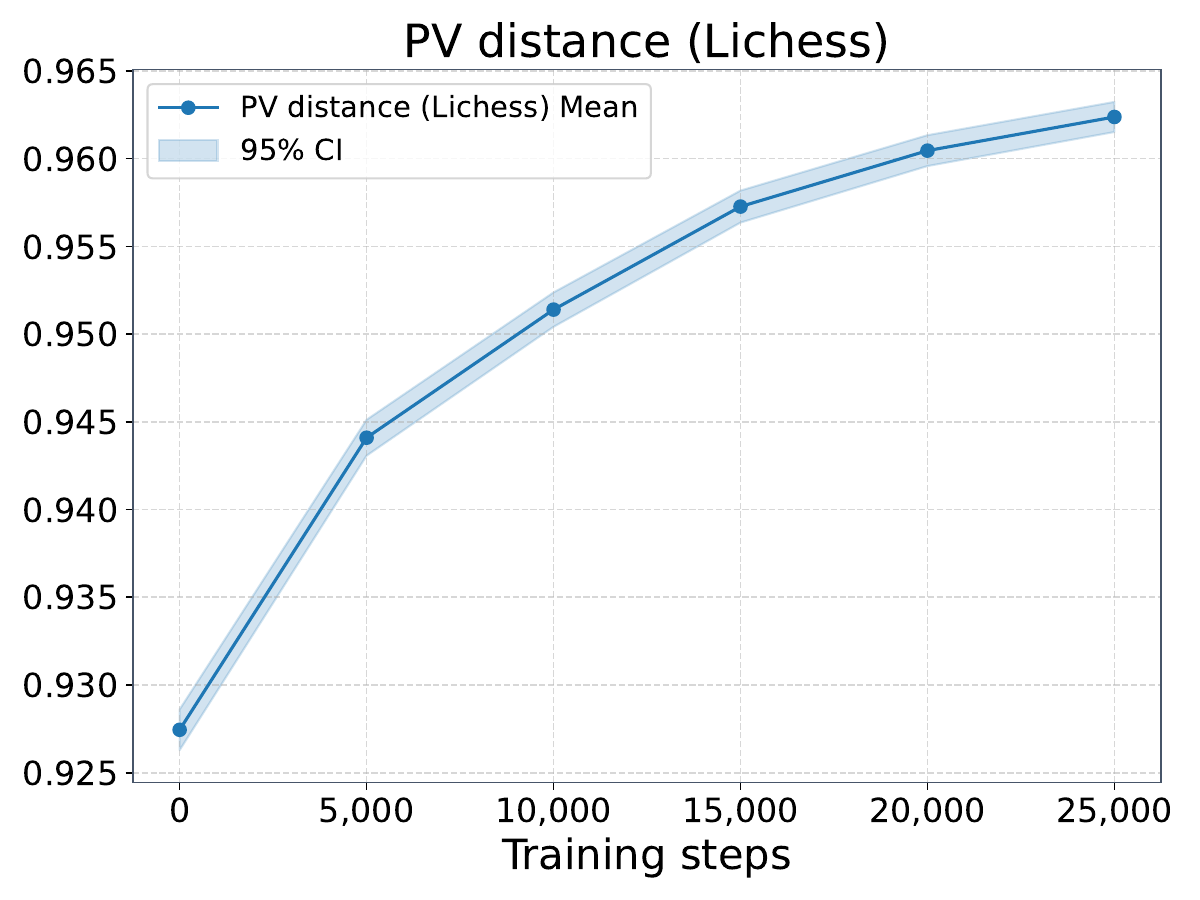}
        \caption{PV distance (Lichess)}
        \label{fig:rl_lichess_pv}
    \end{subfigure}
    \caption{Training progress of run \vtwo in Table~\ref{tab:metrics} during reinforcement learning.}
    \label{fig:rl_training_progress}
\end{figure*}

For reinforcement learning, we initialize the policy from the supervised checkpoint trained with simultaneous best-move prediction, as it achieved the highest baseline puzzle rate of $0.41\%$. We fine-tune this model using the DDPO framework together with the diversity mechanisms described previously.

We run two experiments, \vone and \vtwo. For \vone, we do not condition the generation on the best move or include it in the reward function, while for \vtwo, we sample tokens for the best move randomly and condition the generation of the position on those tokens. While \vone yields the highest counter-intuitive score among all tested models, conditioning on the best move (\vtwo) turns out to be crucial to reduce the policy collapse during RL, as the model is forced to generate positions with different solutions. We discuss experiment \vone and the differences of the two runs in more detail in Appendix~\ref{sec:other rl run} and from here on focus on the \vtwo results.

Figure~\ref{fig:rl_training_progress} presents the progression of the \vtwo RL run. Figures~\ref{fig:rl_uniqueness}--\ref{fig:rl_move_match} show the evolution of the main performance metrics, while the remaining plots characterize the diversity of the generated positions. The initial improvement in reward is driven primarily by the legal rate. During training, the proportion of illegal positions decreases by an order of magnitude, from around $1.0\%$ to around $0.1\%$. The average counter-intuitiveness score $r_{\mathrm{cnt}}(s)$ also improves slightly if filtered for uniqueness and theme matching. As shown in Figure~\ref{fig:rl_counter_intuitive}, its mean increases slightly from below $0.0075$ to above $0.0085$. However, as shown in Table~\ref{tab:metrics}, the proportion of positions with unique solutions that pass the counter-intuitiveness check decreases. The increase in the average counter-intuitiveness together with the lower proportion of positions exceeding the threshold suggests that the distribution of counter-intuitiveness scores becomes more concentrated during training.

Another metric that improves substantially is the uniqueness rate, which increases from below $45\%$ to above $65\%$. The theme-match rate also increases significantly. Since one of our main goals is to improve the controllability of generation, these gains in uniqueness and theme matching are valuable. The proportion of positions that pass both the uniqueness and theme criteria increases from 22.37\% after supervised training to 42.32\% after RL. The model generates more positions with unique solutions that match the requested tactical themes, even though fewer of them are sufficiently difficult according to the counter-intuitiveness criterion.

Figures~\ref{fig:rl_self_board}--\ref{fig:rl_lichess_pv} show the distance metrics of the generated positions over the course of training. All four metrics exhibit an overall upward trend, although the board distances also display a clear periodic pattern, the cause of which remains unclear. We select the checkpoint at 20,000 training steps based primarily on the PV self-distance, which we consider to be the most important diversity metric for practical use. Although this metric increases throughout most of training, it decreases slightly at the final checkpoint. We therefore select the checkpoint at 20,000 steps and stop training after 25,000 steps despite no real signs of model collapse have been observed.

Table~\ref{tab:metrics} presents the final puzzle-quality metrics for the RL model at step 20,000 for version two and at step 3,500 for version one. We are able to train version two for much longer, as we prevent policy collapse by conditioning on the best move. Consistent with the training curves of version two, RL improves the legal, uniqueness, and theme-match rates, while the counter-intuitiveness and overall puzzle rates decrease. This result differs from that of \citet{feng2025generatingcreativechesspuzzles}, whose setup primarily improves the counter-intuitiveness rate while leaving uniqueness nearly unchanged. Their results are much closer to our version one, which we describe in more detail in Section~\ref{sec:other rl run}. The differences between \citet{feng2025generatingcreativechesspuzzles} and our runs likely arise from our conditioning scheme and reward formulation, reflecting the different objectives of the two models. Whereas \citet{feng2025generatingcreativechesspuzzles} focus on increasing the yield of difficult and creative puzzles, our primary objective is to improve the controllability of the generation. Finally, Table~\ref{tab:distances} shows that RL also slightly improves the novelty of the generated positions. Moreover, both the board and PV distances increase relative to the supervised model, both when measured against other generated positions and against the Lichess dataset.

\subsection{Manual control via conditioning}

\begin{figure*}[t]
    \centering
    \includegraphics[width=\textwidth]{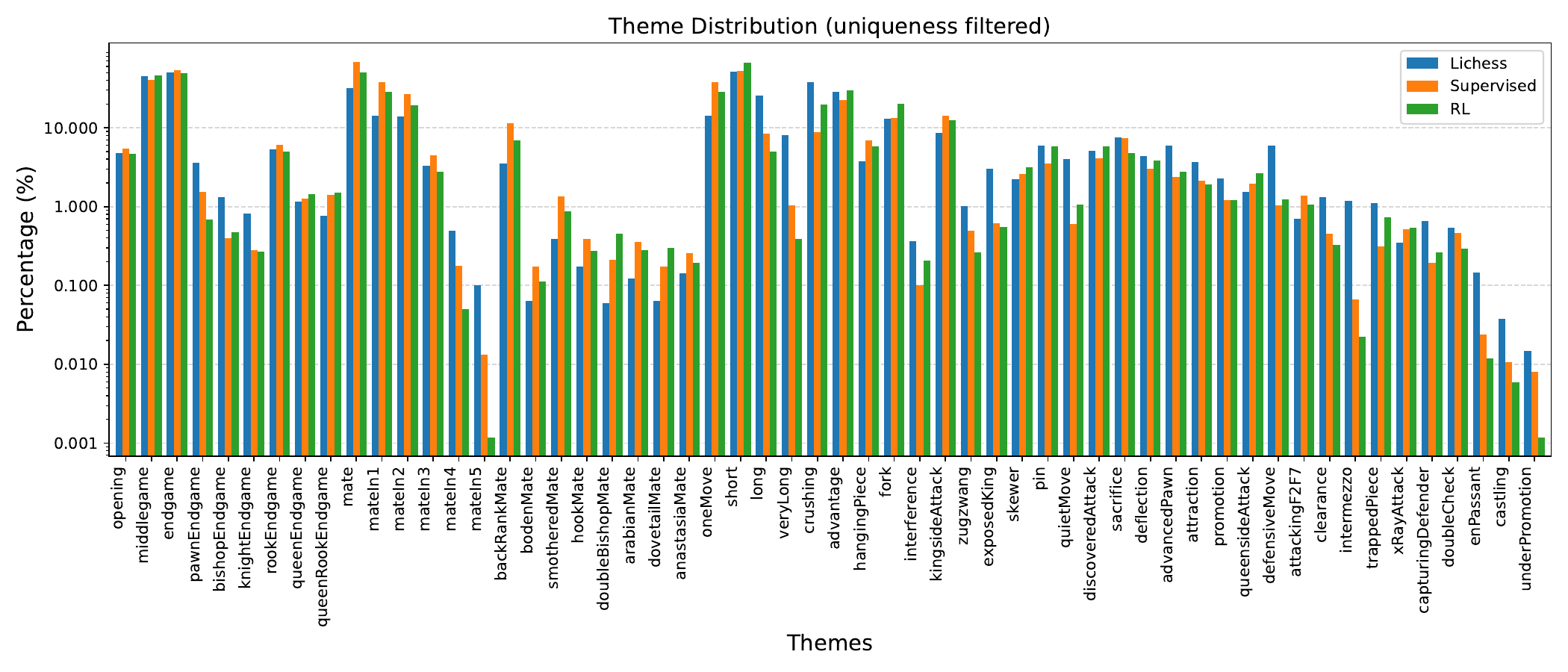}
    \caption{The distribution of themes for the Lichess puzzles dataset, our supervised model and the \vone RL model. All positions are first filtered for positions with unique solutions and the percentages are computed across the filtered dataset.}
    \label{fig:theme_dist}
\end{figure*}

A key contribution of our conditional generation approach is that it enables manual control over the generated puzzles. While unconditioned models, such as that of \citet{feng2025generatingcreativechesspuzzles}, generate creative puzzles at random, they do not allow users to specify which tactics they want to practice. By conditioning our model on specific themes, we make the generator practically more useful, allowing users to generate targeted practice material for positions they struggle with.

Furthermore, conditioning the generation process is crucial for retaining the full diversity of chess tactics post-reinforcement learning. Because short or simple tactics (such as \textit{one-move} or \textit{mate-in-one} puzzles) are almost never counter-intuitive, a global reward naturally suppresses them in favor of longer, more complex configurations. In contrast, our RL setup enforces theme matching via the reward function \eqref{eq:reward function}, ensuring that the model is only rewarded if the generated puzzle satisfies the requested conditioning. Additionally, the advantages are computed over groups of same themes, which forces the policy to maximize puzzle quality within the context of the requested theme, rather than globally converging to the more difficult themes.

Figure~\ref{fig:theme_dist} illustrates the distribution of tactical themes in the original Lichess dataset, our supervised baseline, and the version one RL model. Both of our models successfully preserve the full spectrum of themes. In particular, themes like \textit{mate-in-one} are successfully retained, despite the fact that they are practically never counter-intuitive.

A key advantage of masked diffusion models compared to autoregressive models is that generation can be done in any order. This property allows one to condition the generative process on any partial token sequences, where only some parts of the sequence are masked. In particular, for our use case, we can select some parts of the board which we want to be present in the final puzzle and generate the remaining board. This would not be possible with an autoregressive models natively,\footnote{It could be possible to achieve similar results with autoregressive model by augmenting the training data with partial board prompts that the model is trained to follow. However, this would require additional training.} as the order of generation is decided beforehand.

\begin{figure*}[t]
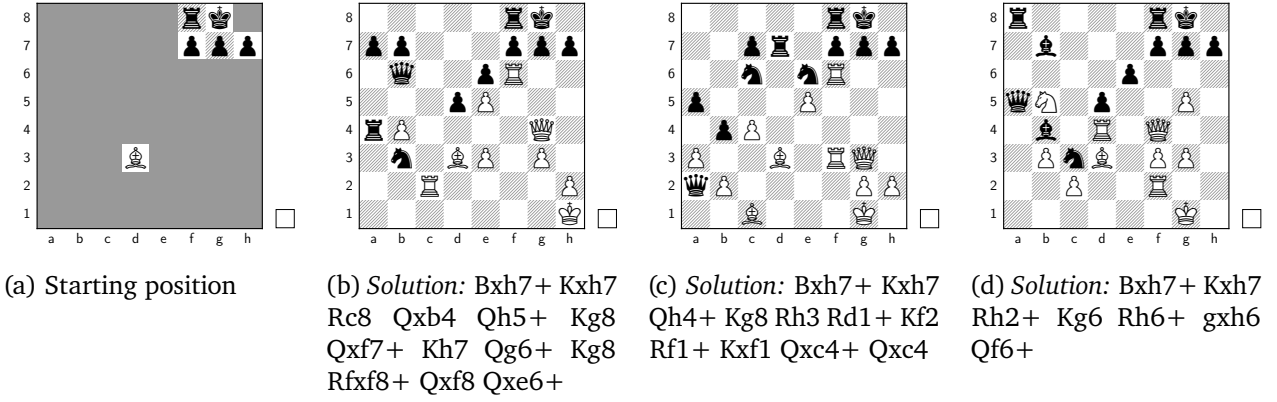

    \begin{subfigure}[t]{0.23\textwidth}
        \centering
        \resizebox{\textwidth}{!}{
            \chessboard[
                setfen=1r3rk1/p1p2ppp/1p1qpn2/4N3/3P1P2/P1PB4/2P5/2K3RQ w - - 4 24,
                pgfstyle=masktoken,
                markfields={a8, b8, c8, d8, e8, h8, a7, b7, c7, d7, e7, a6, b6, c6, d6, e6, f6, g6, h6, a5, b5, c5, d5, e5, f5, g5, h5, a4, b4, c4, d4, e4, f4, g4, h4, a3, b3, c3, e3, f3, g3, h3, a2, b2, c2, d2, e2, f2, g2, h2, a1, b1, c1, d1, e1, f1, g1, h1},
            ]
        }
        \caption{Starting position}
        \label{fig:conditional pos}
    \end{subfigure}
    \hfill
    \begin{subfigure}[t]{0.23\textwidth}
        \centering
        \resizebox{\textwidth}{!}{
            \chessboard[
                setfen=5rk1/pp3ppp/1q2pR2/3pP3/rP4Q1/1n1BP1P1/2R4P/7K w - - 11 25
            ]
        }
        \caption{\textit{Solution:} Bxh7+ Kxh7 Rc8 Qxb4 Qh5+ Kg8 Qxf7+ Kh7 Qg6+ Kg8 Rfxf8+ Qxf8 Qxe6+}
        \label{fig:conditional pos1}
    \end{subfigure}
    \hfill
    \begin{subfigure}[t]{0.23\textwidth}
        \centering
        \resizebox{\textwidth}{!}{
            \chessboard[
                setfen=5rk1/2pr1ppp/2n1nR2/p3P3/1pP5/P2B1RQ1/qP4PP/2B3K1 w - - 2 22
            ]
        }
        \caption{\textit{Solution:} Bxh7+ Kxh7 Qh4+ Kg8 Rh3 Rd1+ Kf2 Rf1+ Kxf1 Qxc4+ Qxc4}
        \label{fig:conditional pos2}
    \end{subfigure}
    \hfill
    \begin{subfigure}[t]{0.23\textwidth}
        \centering
        \resizebox{\textwidth}{!}{
            \chessboard[
                setfen=r4rk1/1b3ppp/4p3/qN1p2P1/1b1R1Q2/1PnB1PP1/2P2R2/6K1 w - - 1 26
            ]
        }
        \caption{\textit{Solution:} Bxh7+ Kxh7 Rh2+ Kg6 Rh6+ gxh6 Qf6+}
        \label{fig:conditional pos3}
    \end{subfigure}
    
    \caption{Three positions generated from a common starting point. The only theme sampling was conditioned with is \textit{sacrifice}. The generation was also conditioned on the bishop sacrifice Bxh7+ and that it is white's move next.}
    \label{fig:condition on position}
\end{figure*}

In Figure~\ref{fig:condition on position} we present an example of conditioning on a partial board configuration and the best move. The initial configuration we use for conditioning is presented in Figure~\ref{fig:conditional pos}. The best move is also used for conditioning, forcing the sacrifice of the bishop on h7 to be the best move. We position the black king in the typical position after castling king side. The remaining position is generated by the model. Figures~\ref{fig:conditional pos1}--\ref{fig:conditional pos3} present three different completions from the initial masked position. Although the first move of the solution is the same for each position, the continuation changes depending on the puzzle.

\section{Hierarchical generation}

\begin{figure*}[t]
    \centering
    \resizebox{\textwidth}{!}{\begin{tikzpicture}[
    level distance=7cm, 
    level 1/.style={sibling distance=16cm},
    level 2/.style={sibling distance=8cm},
    level 3/.style={sibling distance=4cm},
    every node/.style={align=center}
    ]
    \node (L0) [draw, thick, rounded corners, fill=gray!10, inner sep=15pt, text width=10cm] {
        \Huge\textbf{Rook Endgame Generation}\\[2ex]
        \huge Themes: crushing, long, endgame, rook endgame, advanced pawn
    }
    child {node (L1) {\chessboard[smallboard, setfen=6k1/5pp1/3p3p/P1pPr3/1PP5/3RpP2/4rP1P/1R4K1 b - - 1 32, pgfstyle=masktoken, markfields={a8, b8, c8, d8, e8, f8, g8, h8, a7, b7, c7, d7, e7, f7, h7, a6, b6, c6, d6, e6, f6, g6, h6, a5, b5, c5, d5, f5, g5, h5, a4, b4, d4, e4, f4, g4, h4, b3, c3, e3, f3, g3, h3, a2, b2, c2, d2, f2, g2, h2, a1, c1, d1, e1, f1, g1, h1},showmover=False]}
        child {node (L2) {\chessboard[smallboard, setfen=6k1/5pp1/3p3p/P1pPr3/1PP5/3RpP2/4rP1P/1R4K1 b - - 1 32, pgfstyle=masktoken, markfields={a8, b8, c8, d8, e8, f8, g8, a7, b7, c7, d7, f7, h7, a6, b6, c6, d6, e6, f6, g6, b5, d5, f5, g5, h5, a4, b4, e4, f4, g4, h4, b3, c3, e3, g3, h3, a2, b2, c2, d2, f2, g2, h2, c1, d1, e1, f1, g1, h1},showmover=False]}
            child {node (L3) {\chessboard[smallboard, setfen=6k1/5pp1/3p3p/P1pPr3/1PP5/3RpP2/4rP1P/1R4K1 b - - 1 32] \\ \huge exf2+ Kf1 Re1+}}
        }
        child {node {\chessboard[smallboard, setfen=6k1/6pp/p2P4/4rp2/2P5/3RppP1/P3r2P/1R4K1 w - - 1 31,pgfstyle=masktoken,markfields={a8, b8, c8, d8, e8, f8, g8, h8, a7, b7, c7, e7, f7, a6, b6, c6, e6, f6, h6, a5, g5, h5, a4, b4, d4, f4, g4, h4, b3, c3, e3, f3, g3, h3, a2, b2, c2, d2, f2, g2, h2, a1, c1, d1, e1, f1, h1},showmover=True]}
            child {node {\chessboard[smallboard, setfen=6k1/6pp/p2P4/4rp2/2P5/3RppP1/P3r2P/1R4K1 w - - 1 31] \\ \huge d7}}
        }
    }
    child {node {\chessboard[smallboard, setfen=8/4k1p1/2P4p/p1p1p3/1p2P1P1/1PrP1RK1/1r4P1/6R1 w - - 13 51, pgfstyle=masktoken, markfields={a8, b8, c8, d8, e8, f8, g8, h8, a7, b7, c7, d7, e7, f7, g7, h7, a6, b6, d6, e6, f6, g6, h6, a5, b5, c5, d5, e5, f5, g5, h5, a4, c4, d4, e4, f4, g4, a3, b3, d3, e3, f3, g3, h3, a2, c2, d2, f2, h2, a1, b1, c1, d1, e1, h1},showmover=False]}
        child {node {\chessboard[smallboard, setfen=8/4k1p1/2P4p/p1p1p3/1p2P1P1/1PrP1RK1/1r4P1/6R1 w - - 13 51, pgfstyle=masktoken, markfields={a8, b8, c8, d8, e8, f8, g8, h8, b7, d7, e7, f7, g7, h7, a6, b6, d6, f6, g6, h6, b5, c5, d5, e5, f5, g5, h5, c4, d4, e4, f4, a3, b3, d3, e3, g3, h3, a2, d2, h2, a1, c1, d1, h1},showmover=True]}
            child {node {\chessboard[smallboard, setfen=8/4k1p1/2P4p/p1p1p3/1p2P1P1/1PrP1RK1/1r4P1/6R1 w - - 13 51] \\ \huge Rf7+ Kxf7 c7}}
        }
        child {node {\chessboard[smallboard, setfen=8/R6p/1PP3k1/3P2p1/1p2P3/2r4P/1r1p2P1/6RK b - - 9 37,pgfstyle=masktoken,markfields={a8, b8, c8, d8, e8, f8, h8, a7, b7, c7, d7, f7, g7, h7, a6, b6, d6, f6, g6, c5, e5, f5, c4, e4, f4, g4, a3, b3, d3, f3, g3, h3, a2, d2, f2, h2, b1, c1, d1, e1, h1},showmover=False]}
            child {node {\chessboard[smallboard, setfen=8/R6p/1PP3k1/3P2p1/1p2P3/2r4P/1r1p2P1/6RK b - - 9 37] \\ \huge Rb1 Rxb1 Rc1+}}
        }
    }
    child {node {\chessboard[smallboard, setfen=6k1/3R2p1/1pp2P1p/pPp1p3/6rP/P7/1r4P1/3R2K1 w - - 1 33, pgfstyle=masktoken, markfields={a8, b8, d8, f8, g8, h8, a7, c7, d7, e7, f7, g7, h7, a6, c6, d6, e6, f6, h6, a5, b5, c5, d5, f5, g5, h5, d4, e4, f4, h4, a3, b3, d3, e3, f3, g3, h3, a2, c2, d2, e2, f2, g2, h2, a1, b1, c1, e1, f1, g1},showmover=False]}
        child {node {\chessboard[smallboard, setfen=6k1/3R2p1/1pp2P1p/pPp1p3/6rP/P7/1r4P1/3R2K1 w - - 1 33, pgfstyle=masktoken, markfields={a8, b8, d8, g8, h8, a7, c7, d7, e7, f7, h7, a6, c6, d6, f6, a5, b5, c5, d5, g5, h5, d4, e4, f4, h4, d3, e3, f3, g3, h3, c2, d2, e2, f2, g2, a1, b1, c1, e1, f1},showmover=False]}
            child {node {\chessboard[smallboard, setfen=6k1/3R2p1/1pp2P1p/pPp1p3/6rP/P7/1r4P1/3R2K1 w - - 1 33] \\ \huge Rxg7+ Rxg7 Rd8+ Kh7 fxg7}}
        }
        child {node {\chessboard[smallboard, setfen=6k1/p1p4p/1p2P3/2p1p3/6rp/8/Pr3PP1/3R1RK1 w - - 0 36,pgfstyle=masktoken,markfields={a8, b8, d8, f8, a7, c7, d7, e7, f7, g7, h7, a6, c6, d6, e6, f6, b5, c5, d5, f5, h5, d4, e4, f4, h4, a3, b3, d3, e3, f3, g3, h3, c2, d2, e2, f2, h2, a1, b1, c1, g1},showmover=False]}
            child {node {\chessboard[smallboard, setfen=6k1/p1p4p/1p2P3/2p1p3/6rp/8/Pr3PP1/3R1RK1 w - - 0 36] \\ \huge Rd8+}}
        }
    };
    \path (current bounding box.west) -- ++(-1.5cm,0) coordinate (LabelCol);
    \node[anchor=east, font=\Huge\bfseries] at (LabelCol |- L0) {Step 256};
    \node[anchor=east, font=\Huge\bfseries] at (LabelCol |- L1) {Step 213};
    \node[anchor=east, font=\Huge\bfseries] at (LabelCol |- L2) {Step 171};
    \node[anchor=east, font=\Huge\bfseries] at (LabelCol |- L3) {Step 0};
    
    \end{tikzpicture}}
    \caption{The hierarchical generation process of a rook endgame. In predetermined steps, we split the generative process in four and obtain a tree of positions for a single set of themes.}
    \label{fig:hierarchical sampling}
\end{figure*}
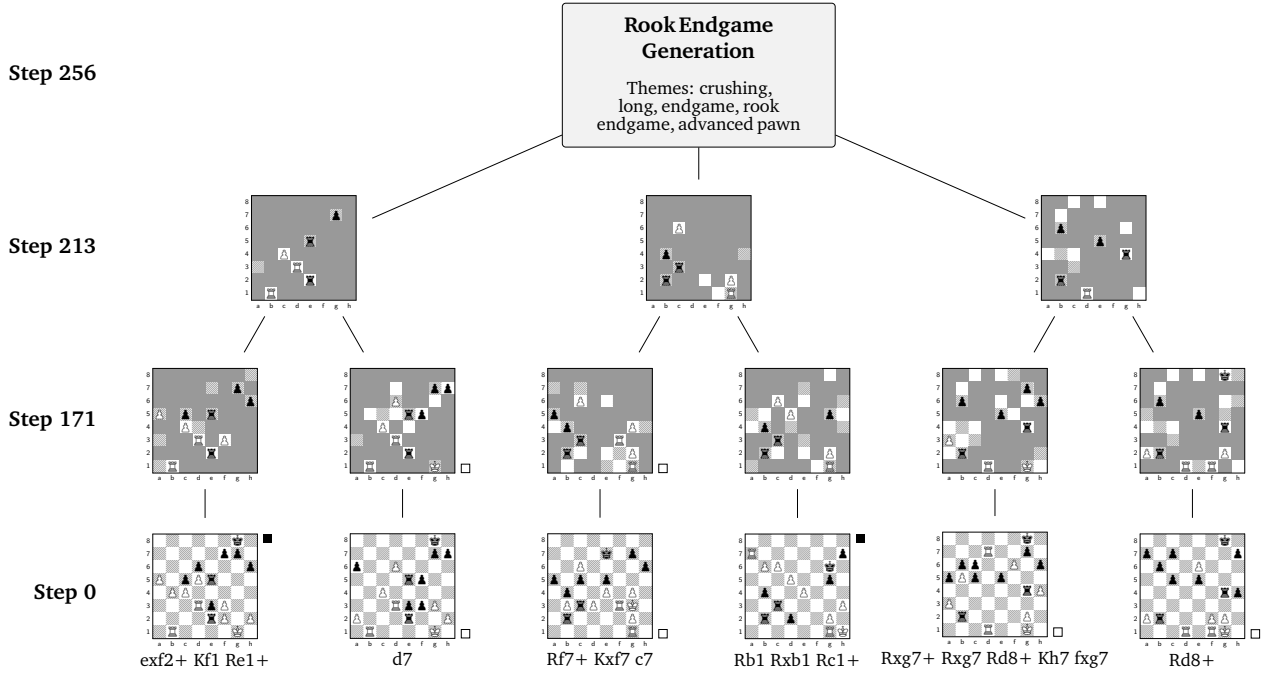

To further experiment with partial board conditioning, we use our model to generate new positions hierarchically. In a standard masked diffusion reverse process, all tokens are unmasked gradually over the total number of time steps to form a single final state. In our hierarchical setup, we instead define a set of specific time steps at which we split the sampling process. We run the denoising process on a single state until a predetermined split step is reached. At this juncture, the partially unmasked state, which still contains masked tokens, is duplicated into multiple independent branches. Each branch then continues its own reverse process. Because the remaining masked tokens are sampled independently in each branch, they resolve into different final piece configurations while retaining the shared tokens that were unmasked prior to the split.

Figure \ref{fig:hierarchical sampling} illustrates how this generative process diverges from a common starting point. After sampling the initial themes, we begin the sampling process by splitting the initial state into four trajectories at step 256. We then apply the reverse process to each branch until the next predetermined split point at step 213, where we branch the generation again. Three of these four initial branches are shown on level two of Figure \ref{fig:hierarchical sampling}. We repeat this alternating procedure of splitting and denoising across specific time steps for a total of six iterations, ultimately ending up with $4^6 = 4096$ positions. From this batch, we select a few interesting candidates to visualize in each subtree. Interestingly, we observe that the majority of the diversity between puzzles is established early in the generative process, while the latter splits mostly end up in similar puzzles. If we look at positions where the latest common ancestor is at step 128, with approximately half of the tokens masked, most puzzles share very similar principal variations. This likely occurs because critical components, such as kings, rooks, and important pawns, have already been unmasked by this stage, leaving minimal room to vary the best move.

This hierarchical generation strategy highlights a structural difference of masked diffusion models over standard autoregressive sequence models. While a similar sampling method can be applied to autoregressive models, a major disadvantage is that the shared components between branches are strictly constrained to a sequential prefix. Because autoregressive models generate tokens in a fixed order, any shared ancestry dictates that the beginning of the string remains identical across all branches, forcing all variation into the later tokens. In the context of chess puzzle generation, this means the generated variations would always share the same pieces on the starting ranks, which will not appear random to a human eye. In contrast, our masked diffusion model unmasks tokens globally based on a masking schedule rather than their position in the sequence. Consequently, the shared tokens of a sub tree can be distributed anywhere on the board, allowing the hierarchical variations to appear naturally random to human observers.

\section{Conclusions and discussion}

We presented a masked diffusion framework for controllable chess puzzle generation, demonstrating that diffusion models are well suited to highly constrained creative tasks. Unlike autoregressive generation, masked diffusion naturally supports conditioning on arbitrary subsets of the output, enabling generation conditioned on partial board positions and best moves, in addition to the tactical themes. We further showed that jointly predicting the best move alongside the board improves generation quality, suggesting that this auxiliary task may encourage richer internal representations of chess concepts rather than merely serving as an additional prediction target. After supervised training, approximately 22.37\% of the generated positions had a unique solution while also matching the requested themes. Reinforcement learning based on Denoising Diffusion Policy Optimization almost doubled this yield to 42.32\% by producing more unique and positionally diverse puzzles while preserving practical controllability. Finally, we release the first open-weight models for conditional chess puzzle generation to the community.

For future work, preventing model collapse remains as the most difficult part to get right. Our random sampling of the best move improves the move diversity, but still puzzles with similar themes exhibit similar ideas despite the main line being different. Addressing this challenge is an important direction for future work. Possible approaches include refining diversity filtering or improving the reward formulation. Beyond reinforcement learning, investigating hybrid procedures that combine diffusion models with iterative search or chain-of-thought reasoning could facilitate the construction of puzzles of even greater complexity. Such systems could also provide more value for LLM systems in general.

More broadly, we believe that the principles explored here extend beyond chess puzzles. Many generation problems require simultaneously satisfying strict structural constraints while producing controllable outputs. The combination of masked diffusion and auxiliary prediction could provide a promising design pattern for other natural language processing domains. In argumentation, for instance, themes could correspond to rhetorical strategies and the best move to a concluding claim. Ultimately, our results suggest that this architecture is a promising foundation for creative generation where controllability, reasoning, and novelty are equally important.

\section*{Acknowledgments}

We would like to thank Nuutti Hyv\"onen for many useful discussions and feedback and Arno Solin for supporting this project.

\bibliography{aaai2027}

\clearpage
\appendix

\setcounter{figure}{0}
\setcounter{table}{0}

\renewcommand{\thefigure}{A\arabic{figure}}
\renewcommand{\thetable}{A\arabic{table}}

\section{Reproducibility}

\begin{table}[H]
    \captionof{table}{All hyperparameters of our training setup.}
    \label{tab:hyperparameters}

    \small
    
    \begin{tabular}{p{0.45\columnwidth} p{0.45\columnwidth}}
        \toprule
        Parameter & value\\
        \midrule
        \multicolumn{2}{c}{Model}\\
        Parameter count & 268M\\
        Transformer block count & 16\\
        Attention head count & 8\\
        Embedding dimension & 1024\\
        Activation & SwiGLU (widening factor 2.66)\\
        Sequence length & 81 (or 76 with no best move prediction)\\
        Masking schedule & linear\\
        \midrule
        \multicolumn{2}{c}{Tokenizer}\\
        Vocabulary size & 52 (+ 1 mask token)\\
        Theme count & 66 (8 are not checked for in theme criterion)\\
        \midrule
        \multicolumn{2}{c}{Supervised training}\\
        Optimizer & AdamW\\
        Maximum learning rate & $3\cdot 10^{-4}$\\
        Learning rate schedule & linear + cosine annealing\\
        Weight decay & $10^{-4}$\\
        Gradient norm clip & 1.0\\
        Batch size & 1024\\
        Steps & 2,000,000\\
        GPUs & 8x Nvidia H200\\
        \midrule
        \multicolumn{2}{c}{Reinforcement learning}\\
        Optimizer & AdamW\\
        Learning rate & $3\cdot10^{-5}$\\
        $\beta$, $\gamma$, $\epsilon$ & 0.03, 0.03, 0.2\\
        Num groups & 6\\
        Group size & 16\\
        Denoising steps & 64\\
        Temperature & 1.0\\
        Theme distribution & Hand crafted\\
        PPO epochs & 1\\
        GPUs & 1x Nvidia H200\\
        FEN dist limit & 6\\
        PV dist limit & 1\\
        \midrule
        \multicolumn{2}{c}{Metrics}\\
        $\tau_{\text{uni}}$, $\tau_{\text{cnt}}$ & 0.5, 0.1\\
        Stockfish version & dev-20260111-eb5a65ae (dev build from just before SF 18)\\
        \bottomrule
    \end{tabular}
\end{table}

In Table~\ref{tab:hyperparameters}, we present the hyperparameters of our model training. For the diffusion model's denoising network, we employ a transformer encoder with 268M parameters. We base our model architecture on the work of \citet{ruoss2024amortizedplanninglargescaletransformers}, who utilize a 270M parameter transformer for chess. We employ 16 transformer blocks with an embedding dimension of 1024. Additionally, we employ 8 attention heads in each attention layer. This base model results in around 200M parameters. The total parameter count is further increased by our conditioning approach, which is implemented via cross attention in each transformer block. The conditioning accounts for the 68M parameters. With this architecture, we train two different models. As we experiment with best move prediction, we train one model that can predict the best move and one that cannot. This results in a minor decrease in parameters in positional embedding, as the sequence length of the model that predicts the best move has a sequence length of 81, whereas, the other model has a sequence length of 76.

As we condition the generative process on puzzle themes, our models input a one-hot encoding of the themes. We choose themes based on the Lichess puzzler project \citep{lichesspuzzler}, which implements conditions for most themes. To represent a chess board in machine-readable form, we use the Forsyth-Edwards notation (FEN) \citep{fennotation}, which we tokenize in a similar manner as \citet{ruoss2024amortizedplanninglargescaletransformers} and \citet{feng2025generatingcreativechesspuzzles}. They pad the FEN into a fixed length sequence of characters, which are converted into tokens. We modify this procedure slightly by using separate tokens to represent different concepts. For instance, a black bishop and black-to-move are given separate tokens in our procedure even though they are both represented by the character b. Additionally, the best move is tokenized in the universal chess interface (UCI) format and concatenated into the FEN for the best move model.

The supervised training is conducted by minimizing the negative evidence lower bound with gradient descent. We employ the AdamW optimizer \citep{loshchilov2018decoupled} and a learning rate schedule, which starts with a linear warmup for 1,000 steps and proceeds with cosine annealing until the 2,000,000 steps have been reached. The linear warmup makes the learning rate go from  $3\cdot 10^{-6}$ to $3\cdot 10^{-4}$ and the cosine annealing takes it from $3\cdot 10^{-4}$ to $3\cdot 10^{-5}$. We use a small weight decay of $10^{-4}$ and clip the gradient norm at $1.0$. The total batch size is 1024 positions. Finally, the training is conducted on a single node of eight Nvidia H200 GPUs.

Lastly, for the RL phase, we continue utilizing the AdamW optimizer but with a reduced learning rate of $3\cdot10^{-5}$, executing the training on a single Nvidia H200 GPU. During generation, we sample the themes from a theme distribution written by ourselves. We also condition the generation on partially masked moves, which are similarly sampled from a distribution written by ourselves. During the generation of the trajectories, we employ 64 denoising steps and a temperature of 1.0. To perform the policy updates, we use 6 groups of size 16 to compute the advantages. Diversity and realism are maintained with $\beta = 0.03$ and $\gamma = 0.03$. We ensure stable updates with the PPO clip parameter $\epsilon = 0.2$. The reward function evaluates the generated positions using the latest pre-release development build of Stockfish (dev-20260111-eb5a65ae) at the time of the start of this work. In the reward function, positions are considered too close if their board distance is below 6 or their PV distance is below 1. This means that positions are considered different in terms of the PV distance, if the best moves differ even slightly.

\section{Model weights and training}
\label{sec:model weights and training}

One can access the weights of the model at \href{https://huggingface.co/naapeli/chess-puzzle-generator}{Huggingface} and the training code at \href{https://github.com/naapeli/Chess-Puzzle-Generation}{GitHub}.

\section{Theme computation}

\begin{algorithm}[t]
    \caption{Themes match}
    \label{alg:theme match}
    \begin{algorithmic}
        \REQUIRE Requested themes $\mathcal{T}_{\text{req}}$, Generated puzzle themes $\mathcal{T}_{\text{gen}}$
        \REQUIRE Category sets $\mathcal{S}_{\text{state}}$, $\mathcal{S}_{\text{endgame}}$, $\mathcal{S}_{\text{mate\_len}}$, $\mathcal{S}_{\text{mate\_type}}$, $\mathcal{S}_{\text{other}}$
        
        \IF{$(\mathcal{T}_{\text{req}} \cap \mathcal{S}_{\text{state}}) \not\subseteq \mathcal{T}_{\text{gen}}$}
            \RETURN False
        \ENDIF
        
        \IF{\text{``endgame''} $\in \mathcal{T}_{\text{req}}$}
            \IF{$(\mathcal{T}_{\text{req}} \cap \mathcal{S}_{\text{endgame}}) \not\subseteq \mathcal{T}_{\text{gen}}$}
                \RETURN False
            \ENDIF
        \ENDIF
        
        \IF{\text{``mate''} $\in \mathcal{T}_{\text{req}}$}
            \IF{\text{``mate''} $\notin \mathcal{T}_{\text{gen}}$}
                \RETURN False
            \ENDIF
            \IF{$(\mathcal{T}_{\text{req}} \cap \mathcal{S}_{\text{mate\_len}}) \not\subseteq \mathcal{T}_{\text{gen}}$}
                \RETURN False
            \ENDIF
            \IF{$(\mathcal{T}_{\text{req}} \cap \mathcal{S}_{\text{mate\_type}}) \not\subseteq \mathcal{T}_{\text{gen}}$}
                \RETURN False
            \ENDIF
        \ELSE
            \IF{\text{``mate''} $\in \mathcal{T}_{\text{gen}}$}
                \RETURN False
            \ENDIF
        \ENDIF
        
        \IF{$(\mathcal{T}_{\text{req}} \cap \mathcal{S}_{\text{other}}) \not\subseteq \mathcal{T}_{\text{gen}}$}
            \RETURN False
        \ENDIF
        
        \RETURN True
    \end{algorithmic}
\end{algorithm}

\begin{table*}[t]
    \centering
    \renewcommand{\arraystretch}{1.3}
    \begin{tabular}{l p{10cm}}
        \hline
        \textbf{Category} & \textbf{Themes} \\
        \hline
        State-of-game & Opening, Middlegame, Endgame \\
        Type-of-endgame & Pawn endgame, Bishop endgame, Knight endgame, Rook endgame, Queen endgame, Queen rook endgame \\
        Type-of-checkmate & Mate, Back rank mate, Boden mate, Smothered mate, Hook mate, Double bishop mate, Arabian mate, Dovetail mate, Anastasia mate \\
        Length-of-checkmate & Mate in 1, Mate in 2, Mate in 3, Mate in 4, Mate in 5 \\
        Length-of-puzzle & One move, Short, Long, Very long \\
        Winning & Crushing, Advantage \\
        Other & Hanging piece, Fork, Interference, Kingside attack, Zugzwang, Exposed king, Skewer, Pin, Quiet move, Discovered attack, Sacrifice, Deflection, Advanced pawn, Attraction, Promotion, Queenside attack, Defensive move, Attacking f2/f7, Clearance, Intermezzo, Equality, Trapped piece, X-ray attack, Capturing defender, Double check, En passant, Castling, Underpromotion \\
        \hline
    \end{tabular}
    \caption{The complete list of themes supported by the Lichess puzzler pipeline, grouped by their respective categories.}
    \label{tab:themes}
\end{table*}

To identify the tactical themes present in a generated position, we employ the Lichess puzzler pipeline, which iterates over the principal variation to detect specific themes. We group these themes into several distinct categories, which include the following: state of game, type of endgame, type of checkmate, length of checkmate, length of puzzle, winning states, and other tactics. Table \ref{tab:themes} enumerates every specific theme we evaluate during the puzzle evaluation process. While the Lichess puzzle dataset contains additional metadata tags, we exclusively focus on objective, position-based themes. We intentionally discard the other tags, such as ``master vs. master'', as they cannot be verified from the board state alone.

During reinforcement learning and evaluation, we determine whether a generated puzzle successfully matches the requested conditions by comparing the generated theme set $\mathcal{T}_{\text{gen}}$ against the requested theme set $\mathcal{T}_{\text{req}}$. Rather than requiring a strict superset match $\mathcal{T}_{\text{req}} \subseteq \mathcal{T}_{\text{gen}}$, we use the specific groups for the check. The complete theme matching logic is detailed in Algorithm \ref{alg:theme match}.

\section{Conditioning on difficulty ratings}

\begin{figure}[t]
    \begin{subfigure}[t]{0.5\textwidth}
        \centering
        \includegraphics[width=\textwidth]{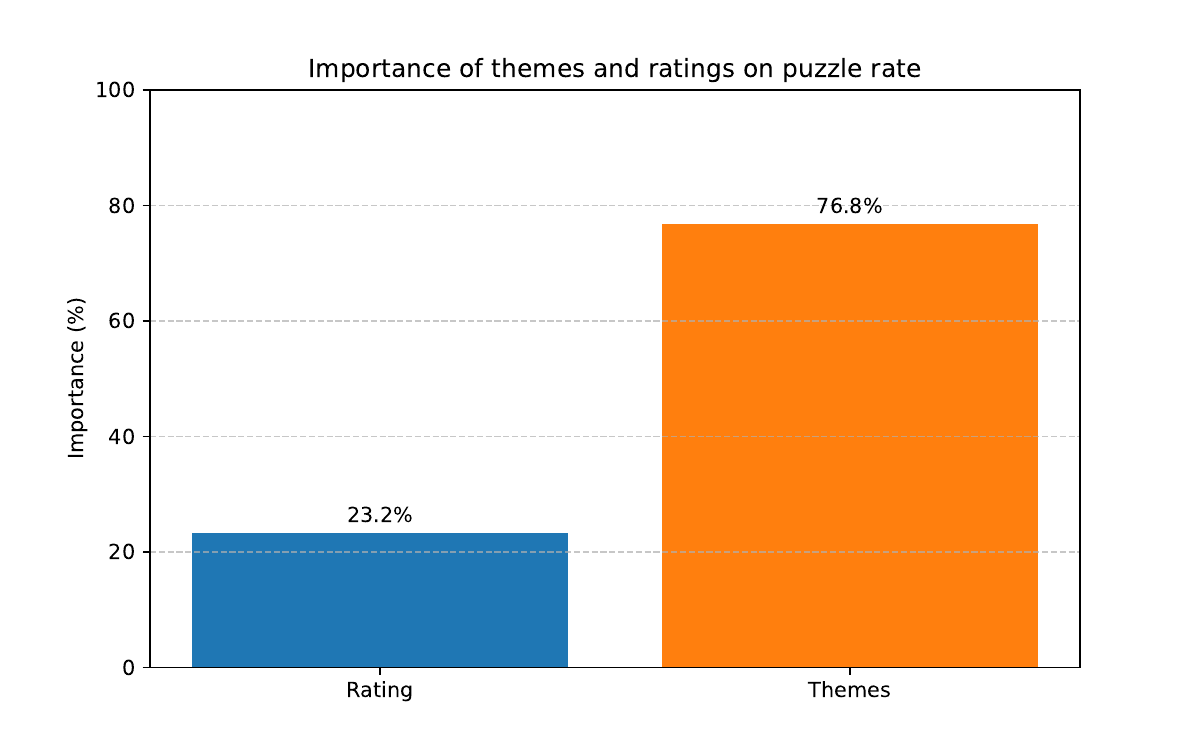}
    \end{subfigure}
    \caption{The impurity based importance of themes and rating on the puzzle rate after supervised training.}
    \label{fig:conditioning importance}
\end{figure}

In addition to the puzzle themes, we condition the generation on puzzle difficulty Elo ratings. We, however, do not observe a substantial effect of the rating on the final puzzle rate compared to the themes. This happens because the themes and the rating are correlated, and the model infers the difficulty from the themes. For instance, a mate-in-one puzzle would almost never have a high difficulty rating in the Lichess dataset, which the model is able to infer from the theme. The effect is further amplified by RL, where the model is rewarded when it generates difficult puzzles with correct themes. The target rating is not part of our reward function, so it is essentially forgotten during RL. In Figure \ref{fig:conditioning importance}, we present the feature importance of the themes and the rating, calculated using a random forest model to predict the puzzle and theme match rate. We observe that after supervised training, the themes explain much more of the puzzle rate uncertainty than the Elo rating.

\section{Reinforcement learning}
\label{sec:other rl run}

\begin{figure*}[t]
    \centering
    \begin{subfigure}[t]{0.24\textwidth}
        \centering
        \includegraphics[width=\textwidth]{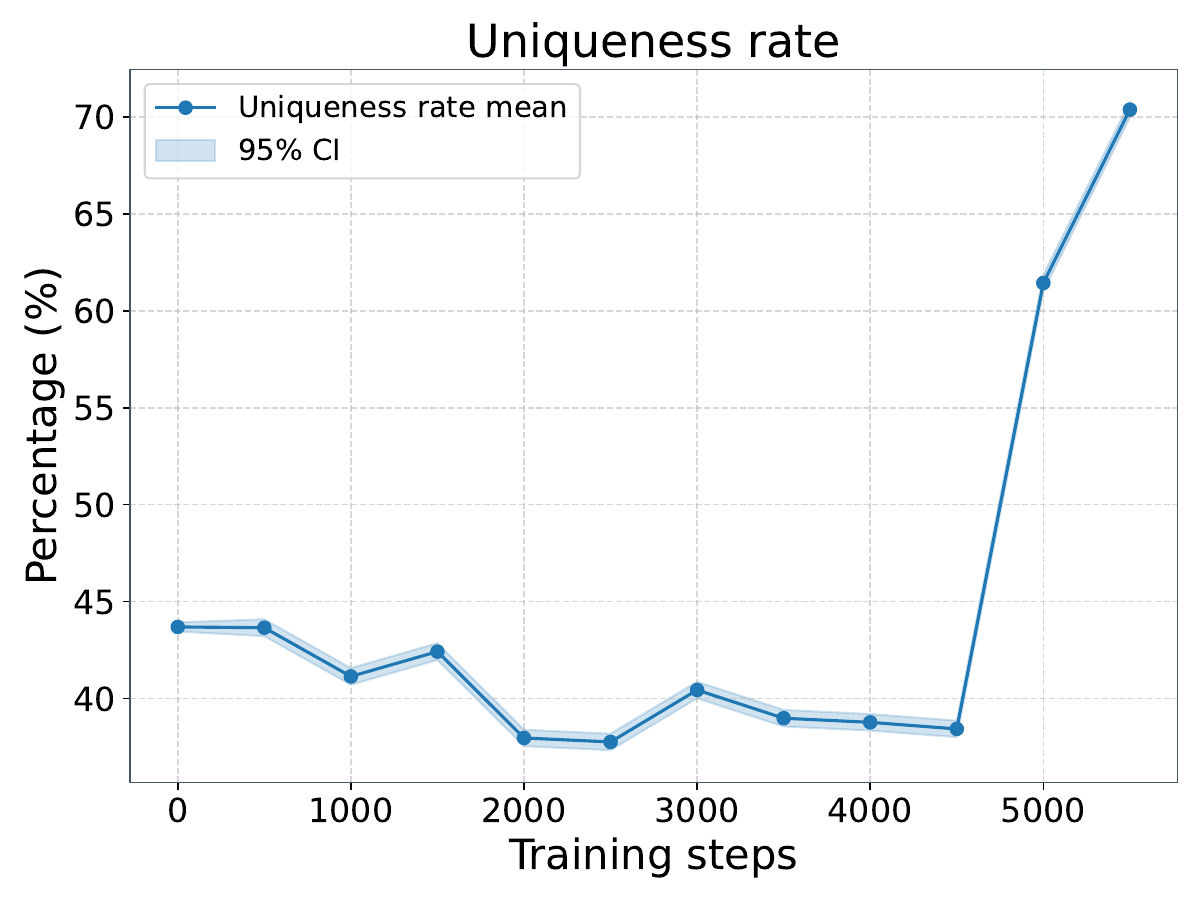}
        \caption{Uniqueness rate}
        \label{fig:rl_uniqueness_v1}
    \end{subfigure}
    \hfill
    \begin{subfigure}[t]{0.24\textwidth}
        \centering
        \includegraphics[width=\textwidth]{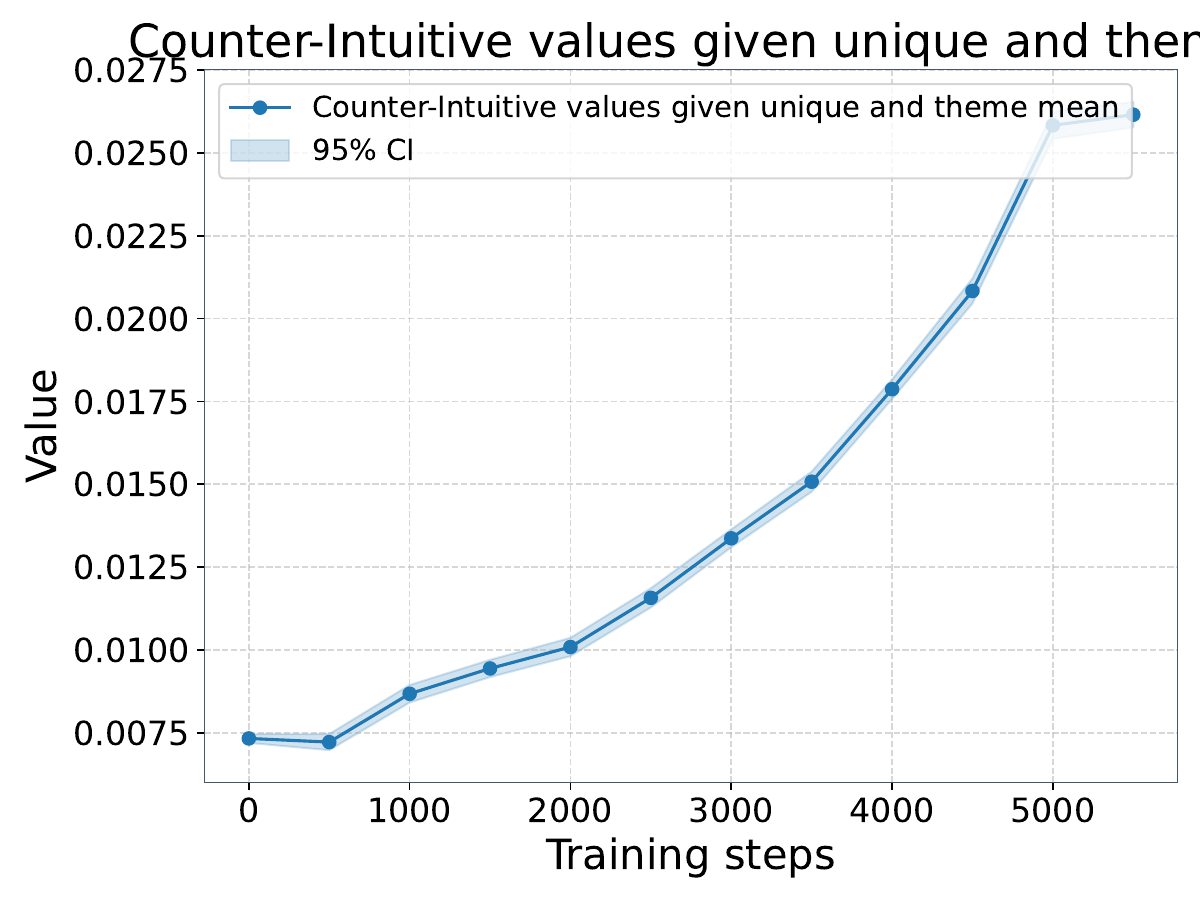}
        \caption{Counter-intuitive values}
        \label{fig:rl_counter_intuitive_v1}
    \end{subfigure}
    \hfill
    \begin{subfigure}[t]{0.24\textwidth}
        \centering
        \includegraphics[width=\textwidth]{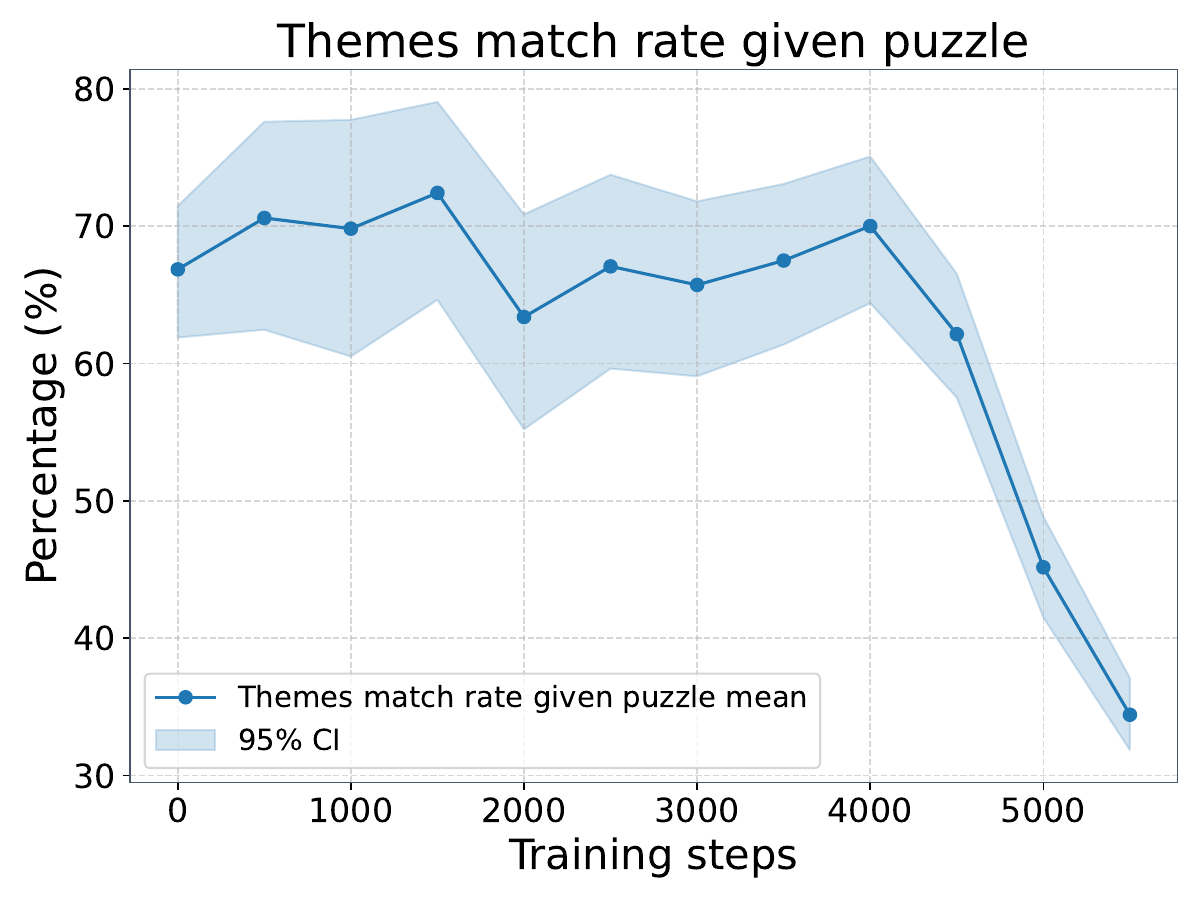}
        \caption{Themes match rate}
        \label{fig:rl_themes_v1}
    \end{subfigure}
    \hfill
    \begin{subfigure}[t]{0.24\textwidth}
        \centering
        \includegraphics[width=\textwidth]{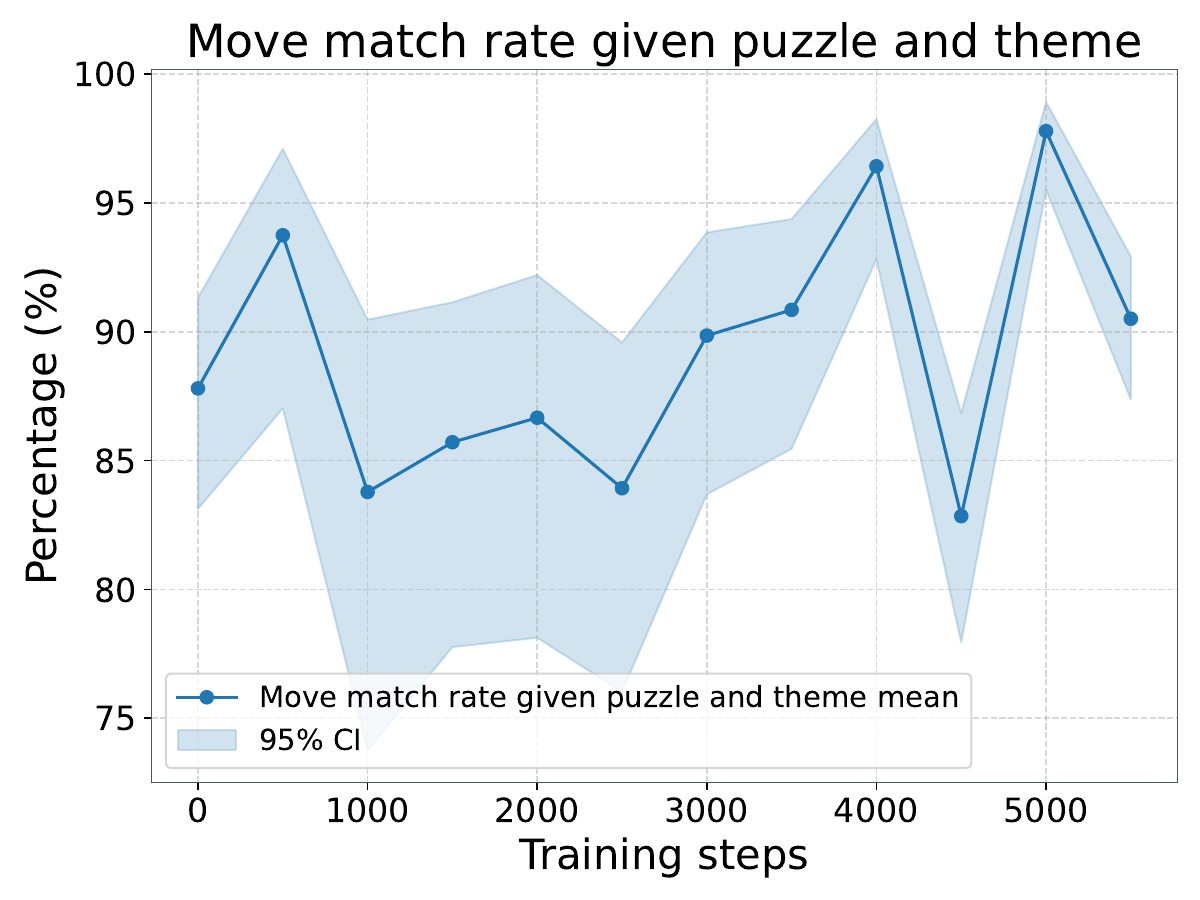}
        \caption{Move match rate}
        \label{fig:rl_move_match_v1}
    \end{subfigure}

    \vspace{0.2cm}

    \begin{subfigure}[t]{0.24\textwidth}
        \centering
        \includegraphics[width=\textwidth]{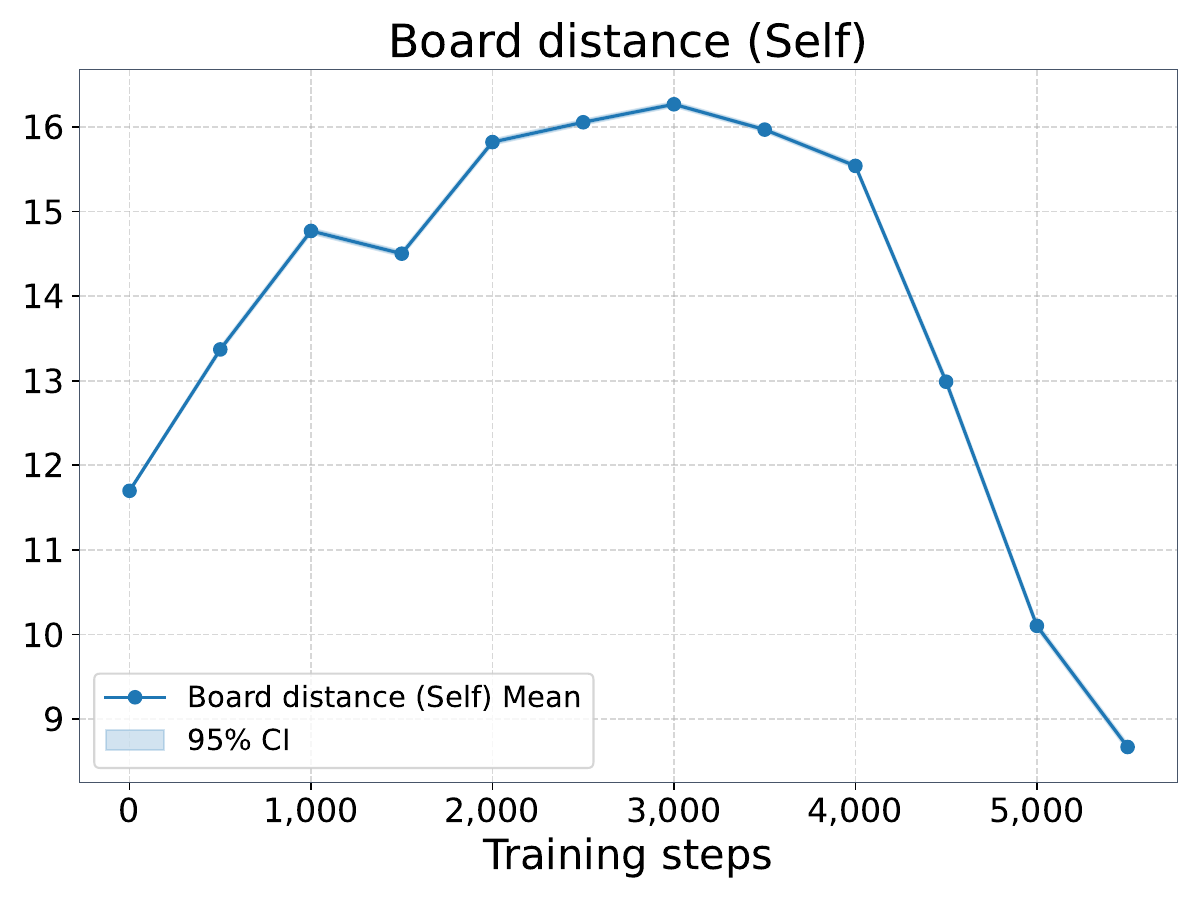}
        \caption{Board distance (Self)}
        \label{fig:rl_self_board_v1}
    \end{subfigure}
    \hfill
    \begin{subfigure}[t]{0.24\textwidth}
        \centering
        \includegraphics[width=\textwidth]{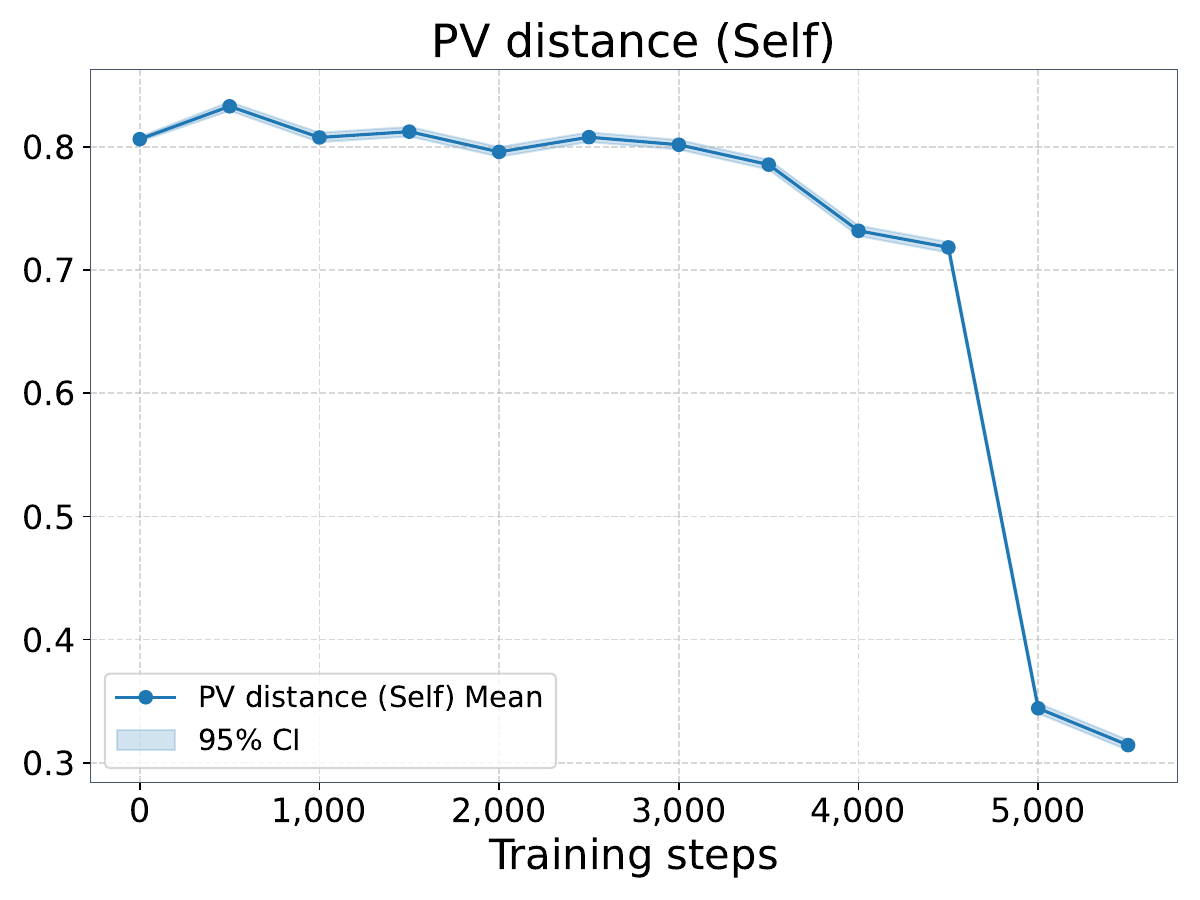}
        \caption{PV distance (Self)}
        \label{fig:rl_self_pv_v1}
    \end{subfigure}
    \hfill
    \begin{subfigure}[t]{0.24\textwidth}
        \centering
        \includegraphics[width=\textwidth]{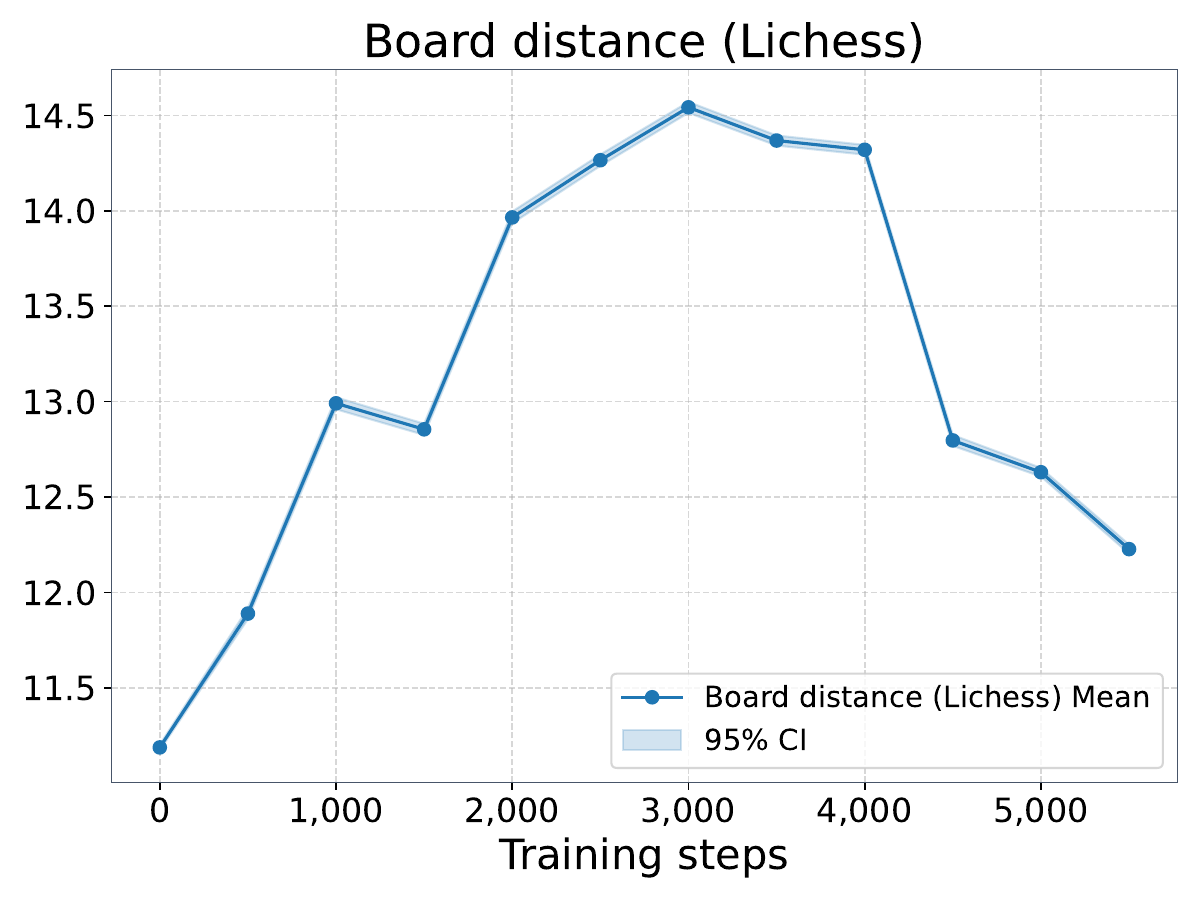}
        \caption{Board distance (Lichess)}
        \label{fig:rl_lichess_board_v1}
    \end{subfigure}
    \hfill
    \begin{subfigure}[t]{0.24\textwidth}
        \centering
        \includegraphics[width=\textwidth]{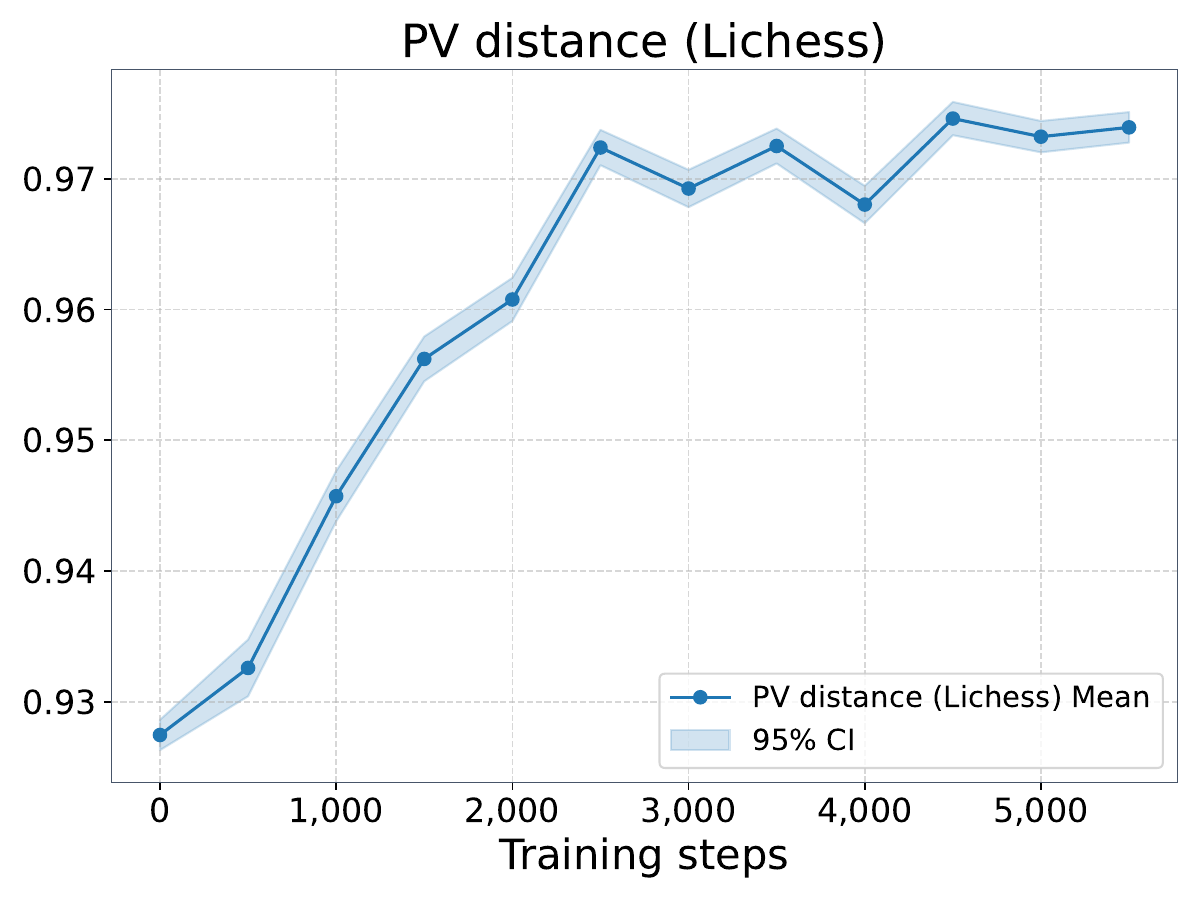}
        \caption{PV distance (Lichess)}
        \label{fig:rl_lichess_pv_v1}
    \end{subfigure}
    \caption{Training progress of run \vone in Table~\ref{tab:metrics} during reinforcement learning.}
    \label{fig:rl_training_progress_v1}
\end{figure*}

While experimenting with reinforcement learning, we observe two interesting runs. The setup presented in Section~\ref{sec:methods} called version two exhibits better controllability due to the higher theme match rates. On the other hand, version one is able to improve the counter-intuitiveness with the cost of lower PV distance. Version one is trained with the following reward function:
\begin{equation}
    \begin{aligned}
        &R(s) = \\
        &\begin{cases}
            -2                                 & \text{if } \neg\mathbb{I}_{\text{legal}}(s)                                                                  \\
            \max(10\cdot r_{\text{cnt}}(s), 0) & \text{if } \mathbb{I}_{\text{legal}}(s) \cdot \mathbb{I}_{\text{filter}}(s) \\
            0                                  & \text{otherwise}
        \end{cases}
    \end{aligned}
\end{equation}
where $\mathbb{I}_{\text{filter}}(s) = \mathbb{I}_{\text{diversity}}(s) \cdot \mathbb{I}_{\text{pieces}}(s) \cdot \mathbb{I}_{\text{theme}}(s) \cdot \mathbb{I}_{\text{uni}}(s)$. Each indicator plays a similar role as in version two. The diversity indicator factorizes as $\mathbb{I}_{\text{diversity}}(s) = \mathbb{I}_{\text{entropy}}(s) \cdot \mathbb{I}_{\text{intra}}(s) \cdot \mathbb{I}_{\text{inter}}(s)$, where $\mathbb{I}_{\text{entropy}}(s)$ filters out low-entropy generations. Notably, the best move condition, as well as the opponents best move diversity filter are missing from version one, whereas the low entropy filter is not used in version two. The training of version one is also conducted by sampling themes from the Lichess dataset for each group, whereas, in version two we sample the themes group wise from a more diverse set of themes. In version two, some tokens of the best move are sampled at random, which improves diversity. Additionally, version one uses much larger KL- and entropy coefficients of 0.1 and 0.2 respectively, which are not needed in version two due to the diversity preserving best move conditioning.

Table~\ref{tab:metrics} and Table~\ref{tab:distances} present the results of the two RL runs. The main differences are related to the distance metrics as well as the counter-intuitive rate. We observe that although the counter-intuitive rate of version one is much higher, the PV distance of version one is lower and affects the model quite a lot in practice. Additionally, the board distances of version one are much higher. This difference results from the higher entropy bonus. For real world usability, we observe that the self PV distance is the most important metric and that version one has a considerably lower diversity in practice making it not as usable as version two despite the higher counter-intuitive rate.

Figure~\ref{fig:rl_training_progress_v1} shows the same results as Table~\ref{tab:metrics} and Table~\ref{tab:distances}. Whereas run \vtwo observes almost constant counter-intuitive values in Figure~\ref{fig:rl_counter_intuitive}, in Figure~\ref{fig:rl_counter_intuitive_v1} we see that the counter-intuitive values increase considerably. On the other hand, the uniqueness in Figure~\ref{fig:rl_uniqueness_v1} and theme match rate in Figure~\ref{fig:rl_themes_v1} decrease apart from the sudden increase in the uniqueness rate at the very end of training resulting from model collapse. We also observe the model collapse in Figure~\ref{fig:rl_self_board_v1} and \ref{fig:rl_lichess_board_v1}, where initially the board distances increase. After 3,000 training steps, the board distances start to decrease rapidly. In Figure~\ref{fig:rl_self_pv_v1}, we see that the PV distance slowly decreases throughout training with a sudden large decrease after 4,500 training steps. Based on the distance metrics, we choose the 3,500th checkpoint as a comparison to run \vtwo.

\end{document}